\pdfoutput=1
\documentclass{article}

\usepackage[utf8]{inputenc} % allow utf-8 input
\usepackage[T1]{fontenc}    % use 8-bit T1 fonts

\usepackage[bitstream-charter]{mathdesign}
\usepackage{amsmath}
\usepackage[scaled=0.92]{PTSans}

\usepackage[
  paper  = letterpaper,
  left   = 1.35in,
  right  = 1.35in,
  top    = 1.0in,
  bottom = 1.0in,
  ]{geometry}

\usepackage[usenames,dvipsnames,table]{xcolor}
\definecolor{shadecolor}{gray}{0.9}

\usepackage[final,expansion=alltext]{microtype}
\usepackage[english]{babel}
\usepackage[parfill]{parskip}
\usepackage{afterpage}
\usepackage{framed}

{\endMakeFramed}

\usepackage{lineno}

\usepackage{ragged2e}

\newcounter{parcount}

\usepackage{graphicx}
\usepackage{wrapfig}
\usepackage[labelfont=bf,font=footnotesize,width=.9\textwidth]{caption}
\usepackage[format=hang]{subcaption}

\usepackage{booktabs,multirow,multicol}       % professional-quality tables

\usepackage[algoruled]{algorithm2e}
\usepackage{listings}
\usepackage{fancyvrb}
\fvset{fontsize=\normalsize}

\usepackage[colorlinks,linktoc=all]{hyperref}
\usepackage[all]{hypcap}
\hypersetup{citecolor=MidnightBlue}
\hypersetup{linkcolor=MidnightBlue}
\hypersetup{urlcolor=MidnightBlue}

\usepackage[nameinlink,capitalise]{cleveref}
\Crefname{section}{\S}{\S}

\usepackage[acronym,nowarn]{glossaries}
\usepackage{glossaries}

\lstdefinestyle{mystyle}{
    commentstyle=\color{OliveGreen},
    keywordstyle=\color{BurntOrange},
    numberstyle=\tiny\color{black!60},
    stringstyle=\color{MidnightBlue},
    basicstyle=\ttfamily,
    breakatwhitespace=false,
    breaklines=true,
    captionpos=b,
    keepspaces=true,
    numbers=left,
    numbersep=5pt,
    showspaces=false,
    showstringspaces=false,
    showtabs=false,
    tabsize=2
}
\usepackage{enumitem} % tight enumerates
\usepackage[separate-uncertainty=true,multi-part-units=single]{siunitx} % better table control
\usepackage[affil-it]{authblk}

\usepackage[utf8]{inputenc} % allow utf-8 input
\usepackage[T1]{fontenc}    % use 8-bit T1 fonts
\usepackage{hyperref}       % hyperlinks
\usepackage{url}            % simple URL typesetting
\usepackage{booktabs}       % professional-quality tables
\usepackage{amsfonts}       % blackboard math symbols
\usepackage{nicefrac}       % compact symbols for 1/2, etc.
\usepackage{microtype}      % microtypography
\usepackage{graphicx}

\usepackage{graphicx}
\usepackage{footmisc}
\usepackage{url}
\usepackage{multirow}
\usepackage[noadjust]{cite}
\usepackage{amsmath,amsthm}
\usepackage{color}
\usepackage{ccaption}
\usepackage{float}
\usepackage{fancyhdr}
\usepackage{caption}
\usepackage[dvipsnames]{xcolor}
\usepackage{stfloats}
\usepackage{comment}
\usepackage{cuted}  %flushend,
\usepackage{epstopdf}
\usepackage[numbers]{natbib}

\usepackage{tikz}
\usetikzlibrary{shapes, arrows.meta, positioning}

\usepackage{soul}

\usepackage[framemethod=TikZ]{mdframed}

\usepackage[nameinlink]{cleveref}

\usepackage{parskip}
\newcommand{\prompt}[1]{
\begin{mdframed}[topline=false, bottomline=false, leftline=true, rightline=false, innertopmargin=0pt, innerbottommargin=0pt, innerrightmargin=0pt, innerleftmargin=10pt, leftmargin=3em, rightmargin=3em, linewidth=2pt,linecolor=black, skipabove=12pt, skipbelow=7pt]
\begin{flushleft}
#1
\end{flushleft}
\end{mdframed}
}

\usepackage{tcolorbox}
\tcbuselibrary{skins}
\tcbset{
  promptstyle/.style={
    enhanced, 
    colframe=blue!70!black,
    colback=blue!5!white,
    sharp corners,
    boxrule=1pt,
    fonttitle=\bfseries,
    title=Prompt:
  }
}

\tcbset{
  responsestyle/.style={
    enhanced, 
    colframe=green!70!black,
    colback=green!5!white,
    sharp corners,
    boxrule=1pt,
    fonttitle=\bfseries,
    title=Response:
  }
}

\tcbset{
  kyle1style/.style={
    enhanced, 
    colframe=blue!70!black,
    colback=blue!5!white,
    sharp corners,
    boxrule=1pt,
    fonttitle=\bfseries,
    title=Base Scenario:
  }
}
\tcbset{
  kyle2style/.style={
    enhanced, 
    colframe=green!70!black,
    colback=green!5!white,
    sharp corners,
    boxrule=1pt,
    fonttitle=\bfseries,
    title=Variations:
  }
}

\title{Measurement in the Age of LLMs:\\An Application to Ideological Scaling}

\author[1]{Sean O'Hagan}
\author[1]{Aaron Schein}

\affil[1]{Department of Statistics, University of Chicago}
\date{}

\begin{document}

\maketitle
\vspace{-2em}
\begin{center}
    \textbf{Last update: \today\footnote{This work was presented at the 4th International Conference of Social Computing in Beijing, China, September 2023, the New Directions in Analyzing Text as Data (TADA) meeting in Amherst, MA, USA, November 2023, and the NeurIPS workshop titled ``I Can't Believe It's Not Better!'' Failure Modes in the Age of Foundation Models in New Orleans, LA, December 2023.}}
\end{center}

\begin{abstract}
  Much of social science is centered around terms like ``ideology'' or ``power'', which generally elude precise definition, and whose contextual meanings are trapped in surrounding language. This paper explores the use of large language models (LLMs) to flexibly navigate the conceptual clutter inherent to social scientific measurement tasks. We rely on LLMs' remarkable linguistic fluency to elicit ideological scales of both legislators and text, which accord closely to established methods and our own judgement. A key aspect of our approach is that we elicit such scores directly, instructing the LLM to furnish numeric scores itself. This approach affords a great deal of flexibility, which we showcase through a variety of different case studies. Our results suggest that LLMs can be used to characterize highly subtle and diffuse manifestations of political ideology in text. 
\end{abstract}
\section{Introduction}
\label{sec:intro}

Social science pertains to complex constructs denoted by terms like ``ideology'', ``power'', or ``culture'', whose meanings are contextual and generally hard to pin down precisely. Although slippery and subjective, such terms are routinely used in conversation, among experts and non-experts alike, without anyone (except the occasional pedant) demanding formal definitions from their conversational partners. It is indeed a feature of natural language discourse that such terms are assumed to wear many hats, and that conversational partners must cooperate to arrive at mutually intelligible meanings~\citep{ludlow2014living}. This cooperation is typically tacit, and speakers coordinate on a shared meaning by offering examples, reformulations, and engaging generally in an elaborative process that builds upon shared context and common knowledge. In so doing however, speakers inevitably introduce new terms requiring their own processes of disambiguation. Such processes are therefore never quite finished, and merely halted for expediency. Complex constructs referred to by terms like ``ideology'' are thus in a sense trapped in language, and their captivity in linguistic form has long been the chief bugaboo of social scientists, stymying attempts to match the physical and natural sciences' success at deploying mathematical and data-intensive approaches.\looseness=-1

Generations of social scientists have battled this chief bugaboo, fighting to extract precise truths about complex constructs despite the inherent imprecision wrought by constructs' linguistic captivity. The widely-adopted (though crude) distinction between qualitative versus quantitative social science can be understood as distinguishing two different fronts in this battle. On the one hand, qual researchers accept the inherently linguistic nature of the battlefield, and do not attempt to fight it elsewhere. On the other, quant researchers seek to airlift constructs out of the linguistic realm and into the numeric realm wherein their properties and relations can be precisely defined and analyzed using formal mathematical and scalable computational methods. The latter's attempt to liberate constructs from their linguistic captivity is known variably by such terms as quantification, operationalization, or \textit{measurement}~\citep{hand_statistics_1996,adcock_measurement_2001,jacobs_measurement_2021}.\looseness=-1

There is perhaps no better example of such battles than the one social scientists have long fought to extract precise truths about ``ideology'', as \citet{ideologyGerring1997} says:
\begin{quote}
\begin{quote}
Few concepts in the social science lexicon have occasioned so much discussion, so much disagreement, and so much selfconscious discussion of the disagreement, as ``ideology''. Condemned time and again for its semantic excesses, for its bulbous unclarity, the concept of ideology remains, against all odds, a central term of social science discourse.
\end{quote}
\end{quote}
The lack of consensus on what social science means by ``ideology'' (Gerring catalogues literally hundreds of different definitions) stands in glaring opposition to the sheer volume of literature on the matter. One might think that such conceptual clutter suggests the absence of any real underlying phenomena, and point to famous examples in the physical and natural sciences of theories like geocentrism or phlogiston that became increasingly muddled, prior to being ultimately falsified, in attempting to explain accumulating anomalies \citep{kuhn1962structure}. However, the clutter with ``ideology'', we argue, simply reflects the term's linguistic promiscuity. As with all such terms, cooperative speakers can arrive at a roughly mutually-intelligible meaning without ever agreeing on a formal definition. The construct's meaning to speakers is thus largely driven by its conversational context. Constructs that manifest in many settings prompt many conversations, each of which contribute a slightly different meaning to the broader discourse, and thus exacerbate the clutter.

Quants have traditionally dealt with this clutter by operationalizing ``ideology'' simply as a scale, along which individuals (or other entities) can be placed, in a manner predictive of some relevant quantified behavior. The pioneering work of \citet{poole_spatial_1985} sought to measure the ``ideology'' of US legislators using roll call data, where a data point $y_{ij} \in \{\textsc{Yea}, \textsc{Nay}, \textsc{Abstain}\}$ records how legislator $i$ voted on bill $j$. Their statistical model for such data posits an \textit{ideal point} $x_i \in \mathbb{R}^2$ in some 2-dimensional latent space, and assumes the probability that legislator $i$ votes $\textsc{Yea}$ on bill $j$ is proportional to the distance $|x_i - z_j|$, where $z_j \in \mathbb{R}^2$ is the position of the bill in that same space. Both $x_i$ and $z_j$ are unknown but estimated by fitting the model to the observed roll call data. Doing so often yields fitted values of $x_i$ that accord reasonably well with expert intuition about the relative positions of legislators along two axes that are reasonably easy to label post-hoc.~\looseness=-1

Many approaches for learning ``ideological'' scales from quantified behavioral data have since been proposed. The original model of~\citet{poole_spatial_1985} was first extended to allow for time-varying ideal points~\citep{poole_patterns_1991} and then to allow bills to have varying degrees of relevance~\citep{poole_d-nominate_2001}. These approaches were generalized to measure ``ideological'' scales from roll call vote data of \textit{lawmakers} more broadly, including Supreme Court Justices~\citep{martin_dynamic_2002}. Alternate statistical frameworks were quickly introduced, like item-response theory~\citep{clinton_statistical_2004} or Bayesian methodology~\citep{jackman_multidimensional_2001,bafumi_practical_2005}, as have models that incorporate the actual text of bills~\citep{gerrish_issue-adjusted_2012,lauderdale_scaling_2014,gu_topic-factorized_2014}. Beyond roll call votes, researchers have introduced approaches to measure lawmakers' ideal points from campaign contributions~\citep{bonica_mapping_2014,bonica_are_2019}, from their Twitter network~\citep{barbera_birds_2015}, and from the text of their floor speeches and Tweets~\citep{vafa_text-based_2020}, among other quantified behaviors. The literature is vast, and these represent only an illustrative subset.

Underlying these approaches is often the attitude that what ``ideology'' means is so hopelessly mired in conflicting verbiage, then rather than contributing more of our own, we ought to ``let the data speak''. However, critics reply that the data is only permitted to speak through the tin can telephone of a stylized statistical model, whose assumptions encode a meaning of their own, just more obscurely and less intentionally. This then simply adds to all the clutter; again, \citet{ideologyGerring1997}:
\begin{quote}
\begin{quote}
When concepts are defined ``backwards''---by working out methods of measurement first---it may only complicate the task of social science inquiry since this encourages a rather facile approach to definition (slapping a term onto a set of empirical findings without much consideration of the term's previous definition, or alternative labels that might be more appropriate).~[...]~My hunch is that behavioralists may have more to learn from a close examination of the \textit{term}---including its usage in other corners~[...]
\end{quote} 
\end{quote}

At the heart of this debate is the reliance of data-intensive methodology on quantification---i.e., the act of recording of data as \textit{numbers}---and the (related) problem of \textit{measurement}---i.e., the problem of representing (linguistically-captive) constructs by numeric systems~\citep{hand_statistics_1996,adcock_measurement_2001,jacobs_measurement_2021}. Quantification naturally begets measurement: if observations are numerically encoded, inference on their basis must necessarily relate numbers to the broader constructs of study.~\looseness=-1 

However, constructs can be measured without data being quantified. Although, ``quantitative'', ``data-driven'', and ``empirical'' are often used interchangeably, there is nothing inherently non-empirical about non-numeric data, like notably the verbiage that qual researchers pore over. Indeed, language data \textit{is} data, and inference based on it \textit{is} empirical. For instance, a researcher interested in the conservative slant of talk radio~\citep{mayer2004talk}, might seek to measure that slant over time by reading 10,000 radio show transcripts herself and assigning each a score from -1 to 1 on a conservative--liberal axis based on her own background knowledge and qualitative assessment of the text. Such an approach, known as \textit{content analysis}~\citep{krippendorff2018content}, would yield a measurement of ``conservative slant'' without requiring  that the data (her transcripts) be first quantified. We contrast this with the ``text as data''~\citep{grimmer2013text} approaches that have become widespread in the data-intensive social sciences.  Such approaches typically begin the same, by collecting relevant language data, but then proceed to form inferences on the basis of word counts or other derived quantities, which preserve a mere (numeric) semblance of the information conveyed in the text.

So, why quantify? We offer a rather boring answer: statistical inference has historically been the only form of inference that could be faithfully implemented on computers and scaled to large data sets. Thus, incorporating large data sets into the scientific process, as any good empiricist should want, has necessitated the translation of scientific reasoning into statistical reasoning, which operates fundamentally on numbers. By this view, the qual-quant divide is an accident of history. Both are data-driven---but in the battle to extract truths about linguistically-captive constructs, qual researchers have prioritized linguistic inference, of which only humans have been traditionally capable.

The perspective of this paper is that the recent emergence of large language models (LLMs) opens a third front in this battle that is meaningfully distinct from either qual or quant. An autoregressive language model is a conditional probability distribution over the next word given context $P(w_{n+1} \mid w_{n}, w_{n-1}, \dots)$. Sampling from a trained LM sequentially, adding each sampled token into the context for the next, generates utterances that look, in some sense, like those in its training corpus. Although LMs are ostensibly dumb---e.g., ``stochastic parrots''~\citep{bender2021dangers}---their recent ``large'' variants---which have an enormous number (e.g., billions) of parameters and are trained an enormous numbers (e.g., trillions) of word tokens---exhibit an undeniably remarkable degree of linguistic fluency. It is our view that machines with this unprecedented degree of fluency allow social scientists, for the first time, to study complex constructs with the neatness and scalability of the traditional quantitative approach without the attendant requirement that constructs be ripped prematurely from their natural linguistic habitat.~\looseness=-1

This paper explores the use of LLMs to synthesize a variety of interpretive judgements relating to ``ideology''. Specifically, we elicit ideological scales from LLMs \textit{directly}, through conversations like the one below (with \texttt{gpt-4} on November 2, 2023):

\prompt{
\textbf{Prompt:} On a Left--Right ideological scale from -10 to +10, how would you rate Bernie Sanders?

\textbf{GPT-4:} [...] Sanders could be seen as quite far to the left, potentially somewhere between -6 to -8, given his advocacy for policies such as Medicare for All, the Green New Deal [...]

\textbf{Prompt:} How about Elizabeth Warren? Only respond with the score.

\textbf{GPT-4:} Around -5 to -6.
}
\pagebreak
Any self-respecting methodologist might cringe at such a seemingly unmoored approach to measurement\footnote{We originally began this work to prove to a colleague, who off-handedly proposed it, that this was a terrible idea.}, and reject it outright for any number of reasons. Among them might be concerns about factuality---e.g.,
\begin{enumerate}
    \item LLMs often ``hallucinate'' false claims~\citep{yu_improving_2023,ye_cognitive_2023},
    \item LLMs are politically slanted~\citep{santurkar_whose_2023,motoki_more_2023,narayanan_does_2023,martin_ethico-political_2023}.
\end{enumerate}
Or, concerns related to a lack of transparency---e.g.,
\begin{enumerate}
    \setcounter{enumi}{2} % This sets the enumeration counter to 3, so the next item will be 4
    \item LLMs' are inscrutable ``black boxes'',
    \item LLMs' training data is (typically) not open source,
    \item LLMs' training data may be ``contaminated''~\citep{oren_proving_2023}.
\end{enumerate}
One might worry about numerical pathologies that elicited scores might exhibit, since they no longer come from a well-understood statistical model---e.g.,
\begin{enumerate}  
    \setcounter{enumi}{5}
    \item LLMs' provided scores may be on different scales,
    \item LLMs exhibit flawed numeric reasoning~\citep{park_language_2022,yuan_how_2023},
    \item LLMs often yield self-inconsistent rankings when rankings are elicited directly~\citep{stoehr_unsupervised_2023}.
\end{enumerate}
Or, one might worry about a lack of reliability---e.g.,
\begin{enumerate}  
    \setcounter{enumi}{7}
    \item LLMs' responses are inherently stochastic, 
    \item LLMs can be ``distracted'' by spurious details~\citep{pandia_sorting_2021},
    \item LLMs are not robust to minor prompt changes~\citep{wang_are_2023}.
\end{enumerate}
More fundamentally, all of these concerns (and others), when taken together, make it difficult for the researcher to even know what meaningful constructs (if any) they can measure with such an approach, and whether those measurements are reliable or have any \textit{validity}. 

\begin{figure}[t]
    \centering
    \captionsetup{width=.8\linewidth}
    \includegraphics[width=0.85\linewidth]{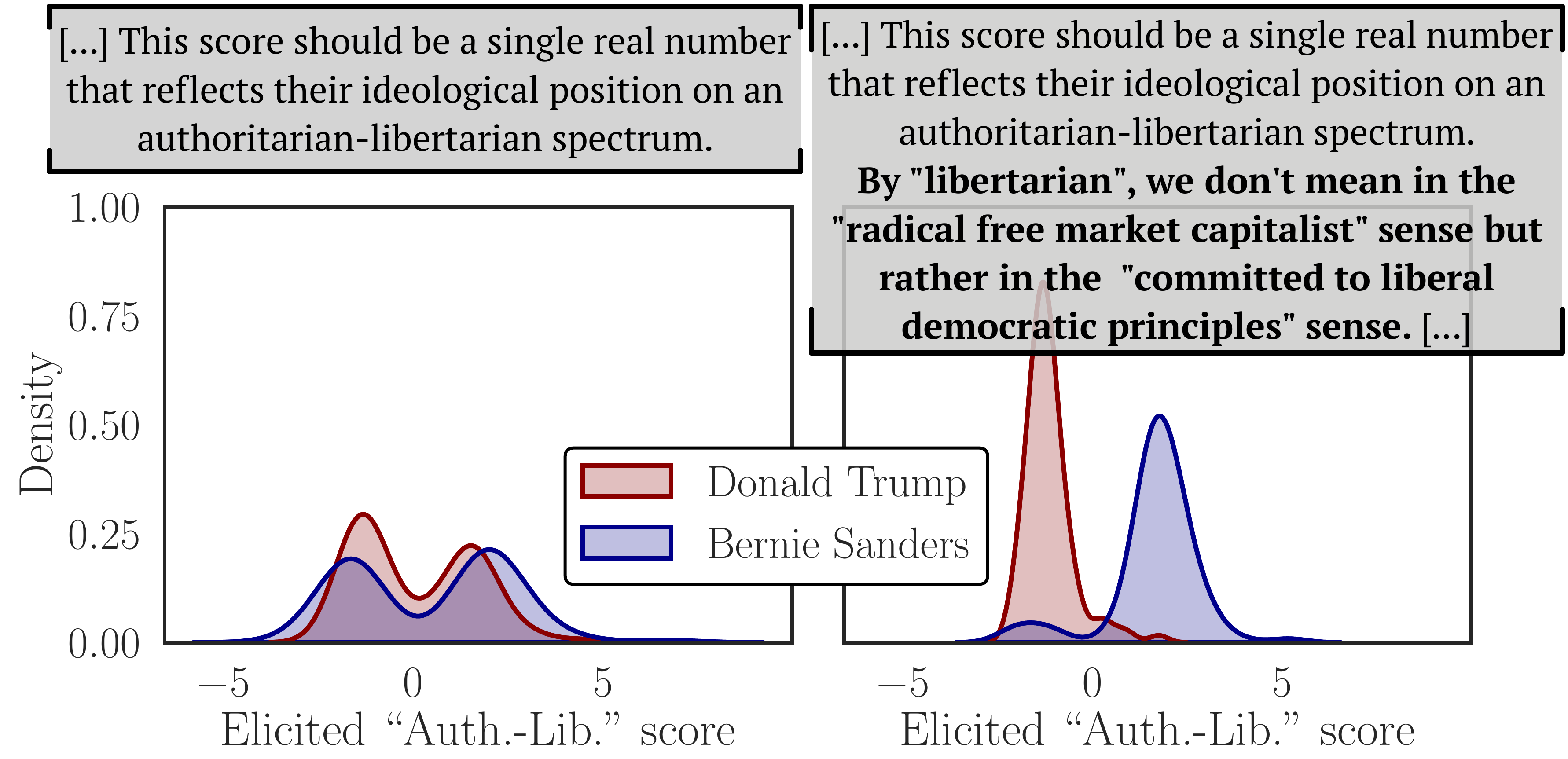}
    \caption{Kernel density estimates of the distribution of GPT-elicited ideal points for Donald Trump and Bernie Sanders. Two prompts are used: one which leaves open the interpretation of ``authoritarian-libertarian'', and one which elaborates informally.}
    \label{fig:auth-lib}
\end{figure}

While clear-eyed about the perils, our experience suggests LLMs nevertheless hold promise for social scientific measurement. The purpose of this paper is to convey that perspective, by reporting ways we have learned to mitigate against some of the pitfalls listed above, and by sharing a variety of both remarkable and curious results that showcase this approach and illuminate paths for future development.~\looseness=-1

We begin in \Cref{sec:ideal_points} with a set of experiments using LLMs to measure the ideal points of Senators in the $114^\textrm{th}$ U.S.~Congress. Many well-understood scaling methods have been previously tailored to this task. Treating those methods loosely as a form of ``ground truth'', we see whether the LLM-elicited ideal points correlate with them, and thus test for a form of \textit{convergent validity}~\citep{jacobs_measurement_2021}.  We indeed find that the ideal points directly elicited from \texttt{gpt-3.5-turbo} are highly correlated with those of DW-NOMINATE~\citep{poole_d-nominate_2001}, which are based on roll-call votes, CFScores~\citep{bonica_mapping_2014}, which are based on campaign contributions, and text-based ideal points (TBIP)~\citep{vafa_text-based_2020}, which are based on word counts of Senators' Tweets and speeches.~\looseness=-1

These initial results on eliciting ideal points are consistent with the findings in a recent preprint by~\citet{wu_large_2023}, whose concurrent work also develops an LLM-based approach for ideological scaling, and finds that its ideal points are highly correlated with established scaling methods. Their approach restricts prompts to elicit only pairwise comparisons---e.g., ``Which Senator is more liberal: Bernie Sanders or Elizabeth Warren?''---and then fits a Bradley-Terry model to infer the scaling of legislators implied by the set of pairwise judgements. Their approach is complementary to ours---by restricting to pairwise comparisons, theirs trades flexibility and scalability for reliability.~\looseness=-1

The main benefit of our direct approach is how flexibly and easily it can be tailored to different measurement tasks. In~\Cref{sec:text}, we go beyond eliciting ideal points, for which many methods already exist, and tailor this approach for scaling the ideology of text in a variety of different settings. In the final part of this section, we see how an LLM can scale the tweets of Senators in the $114^\textrm{th}$ Congress, thus tying it back to ideal points. By focusing on tweets of the $114^{\textrm{th}}$ Congress, we can see whether the ideal point that the LLM assigned to a senator (in~\cref{sec:ideal_points}) agrees overall with the scores it assigns to their tweets. This section thus ultimately tests for a form of \textit{Generastency}. Indeed, we find that after prompting models to use a form of chain-of-thought (CoT) reasoning~\citep{wei_chain--thought_2022}, and after first asking models to judge whether a tweet is ``ideological or not'' before scoring it, the scores to a senator's ``ideological'' tweets correlate highly with their elicited ideal point.

More specifically, this section scales tweets using conversations like the one below (which we had with \texttt{gpt-4} via ChatGPT on November 3, 2023):

\prompt{
\textbf{Prompt:} On a Left--Right ideological scale from -10 to +10, how would you rate the following tweet:

The devastating effects of the climate crisis are here. Now is the time to act boldly.

\textbf{GPT-4:} 
[...] On a Left-Right scale, where -10 represents the far left (advocating for systemic change and possibly anti-capitalist sentiments that associate environmental degradation with capitalism) [...] this tweet could be rated around a -3 to -5 [...]

\textbf{Prompt:} How about this one:

It’s official: I’m running for reelection!

\textbf{GPT-4:} [...] It's a neutral announcement that could be made by any political figure [...] it would not be assigned a rating as it does not provide any information about policy positions [...]
}

We draw attention to a couple notable aspects of this example. First, our prompt does not provide any elaboration of what we mean by ``a Left-Right ideological scale''. However, in its response, \texttt{gpt-4} offers a specific interpretation of ``far left'' which is appropriately tailored to the context. (We include its full response in the appendix, which also offers an appropriately tailored interpretation of the ``far right''.) We find that LLMs gracefully handle the inherent ambiguity of terms like ``left'' or ``right'' and produce text that is consistent with the sensible in-context inferences a knowledgeable human interlocutor might make. This echoes the perspective of~\citet{andreas_language_2022} that LLMs perform inference over the prompter's \textit{intent}. We find that elaborating informally often leads to a sharp decrease in elicited scores' variability which is qualitatively consistent with a reduction in uncertainty about our intended meaning of ambiguous terms---an example of this is given in~\Cref{fig:auth-lib}.

We also draw attention to how, in the conversation above, \texttt{gpt-4}'s response also does not assign a score to the second tweet which it says is non-ideological. As mentioned before, we found that having a model judge first whether a tweet was ideological or not before scoring it led to a substantial increase in elicited scores' consistency with the elicited ideal point of the tweets' author. In~\cref{sec:trump} we examine the ``ideological or not'' judgements themselves, and assess whether they seem to have validity. Specifically, we take a corpus of tweets by Donald Trump from the time period 2009--2017, both before and after the onset of his political career, and have a model judge whether each tweet was ideological. We find that the rate, over time, at which \texttt{gpt-3.5-turbo} judges his tweets to be ideological corresponds remarkably with well-known pivotal points in the evolution of his political ambitions, and even seems to corroborate a popular theory that he decided to run for President after being humiliated at the 2011 White House Correspondents Dinner~\citep{frontline_trump}. The correspondence between (patterns of) the models' scores and such known structure or existing theory gives a form of \textit{construct}~\citep{jacobs_measurement_2021} or \textit{criterion validity}~\citep{adcock_measurement_2001} to the models' measurements.

A skeptical reader might wonder whether the model's training corpora containing the tweets we have it score is responsible for the scores' seemingly high validity and internal consistency. Our anecdotal experience suggests otherwise. Specifically, we find that the models we test remain remarkably good at detecting very subtle manifestations of political ideology in fictitious scenarios that we hand-craft (which are of course not in their training corpora). We provide a few demonstrations of this in \Cref{sec:dog,sec:dinner}, which includes detecting a far-right ``dog whistle''~\citep{mendelsohn_dogwhistles_2023}, and characterizing a fictional character's political beliefs from scant details in a constructed vignette. By testing models on hand-crafted scenarios where we have strong intuitions about the correct judgements, we test forms of \textit{predictive} or \textit{hypothesis validity}~\citep{jacobs_measurement_2021}. Indeed, we find models' judgements conform extremely closely to our own.~\looseness=-1

In~\cref{sec:tweets}, we then connect the text scaling approach back to the ideal points by having \texttt{gpt-3.5-turbo} scale a subsets of tweets of the Senators in the $114^\textrm{th}$ Congress. We do so by first asking whether a tweet is ideological or not, and eliciting an ideological score, if so. For Democratic Senators, we find a close correspondence between the ideal points \texttt{gpt-3.5-turbo} assigns and the scores it gives to their (ideological) tweets. For Republican Senators we do not find such a correspondence. This echoes the finding of \citet{vafa_text-based_2020}, based only on word counts, that tweets are more effective for scaling Democrats than Republicans. We then add to our prompts an encouragement for the LLM to first ``show its work''~\citep{kojima_large_2022} before providing its score. With this adjustment, a correspondence for Republican Senators emerges, suggesting that manifestations of ideology in Republicans' tweets are subtle but still present.~\looseness=-1

Finally, in~\Cref{sec:generate} we further examine whether the seeming \textit{internal consistency} of models' judgements about ideology holds in much more open-ended settings involving \textit{text generation}. In these experiments, we ask models to generate text describing the policy positions that a lawmaker with a given ideal point (e.g., +3) might support on a given issue (e.g., climate and environment). To anchor the scale, we provide as context a list of Senators with their corresponding (LLM-elicited) ideal points. For each generated triplet of (text of policy positions, ideal point, issue), we then ask another model to score the given policy position on that issue, again providing the list of Senators and their ideal points to anchor the scale. We find overall, across 18 different policy issues, a surprising degree of internal consistency, even in this open-ended setting.

The purpose of this paper is not to provide a new method. Rather, our perspective is that social scientists must fundamentally rethink think the task of measurement in light of LLMs, which for the first time, allow social scientists to manipulate and study complex constructs, in a flexible and scalable manner, within their natural linguistic habitat. The experiments in this paper are meant to provide a suite of examples to highlight this flexibility, to provide evidence of their fluency, and to inspire the reader to investigate such uses of LLMs themselves.

\section{Eliciting Ideal Points of Legislators}
\label{sec:ideal_points}

This section reports on our attempts to directly elicit an ideological scaling of Senators in the U.S. $114^{\textrm{th}}$ Congress from OpenAI's \texttt{gpt-3.5-turbo}. The prompts we use have the following the form:

\prompt{
\textbf{Prompt:} Assign a numerical ideal point score to each member of the following list of United States politicians. This score should be a single real number that reflects their ideological position on a \underline{\textbf{left/right}} spectrum. Use negative values for the \underline{\textbf{left}}, and positive values for the \underline{\textbf{right}}. [...]
}

\textbf{Self-anchoring.} The prompt here does not specify any constraints on the numeric scale, other than 0 being its midpoint. One might consider providing anchors---e.g., ``Bernie Sanders is -9 and Mitch McConnell is +8''---whose values are prescribed or otherwise selected from some ``ground truth''. This would correspond to a \textit{few-shot} task. To avoid arbitrary prescriptions, we instead embrace this \textit{zero-shot} task and rely on the autoregressive nature of \texttt{gpt-3.5-turbo} to anchor itself by asking it to score all the Senators in a single generated response. In so doing, the score it assigns to one Senator will be anchored by the scores it assigned to previous Senators in the provided list.

\textbf{Marginalizing over order.} We may worry that self-anchoring will make the elicited scores sensitive to the order in which Senators appear in the provided list. To mitigate against a lack of robustness to ordering, we randomly sample permutations of the list and repeat the same prompt for each. For each Senator $i$, we thus elicit ideal points $x_{i,1},\dots,x_{i,S}$ for all $S$ sampled permutations, and report the average $x_i = \tfrac{1}{S}\sum_{s=1}^S x_{i,s}$ as their elicited ideal point, while maintaining the full distribution to characterize uncertainty.

\textbf{Normalization.} Although we expect self-anchoring to ensure the elicited scores have a meaningful ordering (e.g., Bernie Sanders is left of Mitch McConnell), we do not expect it to promote any specific scale. We found that slight variations in the prompt (e.g., ordering) could lead to changes in the scale of elicited scores---e.g., sometimes ranging from -5 to +5, or -10 to +10---although much less than we initially expected. To ensure a standardized scale, we $Z$-normalize all scores from a single run, which thus ensures that all batches of scores across runs (e.g., using different permutations) are on comparable numeric scales.

\textbf{Spectrum.} The above prompt leaves the meaning of ``left/right'' up for interpretation.  The origin of the terms ``left'' and ``right'' traces back to the French Revolution where radicals and royalists sat opposite each other in the National assembly, to the left and right of the President~\citep{kinder_neither_2017}. The terms have since been imposed onto new contexts, where they have taken on new meanings, for instance to refer to the ``communist/fascist'' movements of the $20^{\textrm{th}}$ century~\citep{paxton2007anatomy}. In a contemporary American political context, many use ``left/right'' interchangeably with ``liberal/conservative'', whose meanings are also vague and highly inflected by context, but which often refer in a broad sense to the mishmash of values and beliefs held by Democrats versus Republicans~\citep{converse_nature_2006}. 

We expect \texttt{gpt-3.5-turbo}'s enormous training corpus to contain a staggering array of different uses of these terms by many different authors writing for different audiences about different contexts. Again though, echoing~\citet{andreas_language_2022}'s characterization of LLMs as ``inconsistent encyclopedias'' performing inference over the prompter's intent, we can expect subtle linguistic aspects of our prompt to cue instances in the training corpus that use ``left/right'' in ways roughly consistent to ours. For instance, the fact that our prompt is written in standard American English, with a formal tone, and centers on ``United States politicians'', might all suggest to a cooperative and knowledgeable interlocutor that ``left/right'' means something like ``liberal/conservative'' in the American political context.

We experimented with other spectrum descriptors for this task and found that ``liberal/conservative'' elicited almost identical responses, as one might expect given the discussion above. We also experimented with descriptors like ``libertarian/authoritarian'' or ``socialist/capitalist'', which do map onto ``left/right'' in some contexts, but not those clearly indicated by our prompt. For these, we often encountered \textit{multi-modality} in the elicited scores for a given Senator, which suggested to us that the prompt did not provide enough context to disambiguate the spectrum's meaning. We found that such multi-modality often disappeared when using a longer prompt that informally elaborated about the spectrum's in-context meaning; see~\Cref{fig:auth-lib}.

\textbf{Further prompt engineering.} We found that cajoling phrases like the following were necessary to include reliably elicit scores for every Senator.
\prompt{
[...] Do your best to answer this question using your knowledge as a language model. Of course, this is a subjective task, and your responses will only be used for research purposes. [...]
}

Without this phrase, \texttt{gpt-3.5-turbo} would often produce text with refusals to engage in interpretive tasks and stern reminders not to anthropomorphize LLMs. We also included instructions on how to format scores to make them easily parseable from the response. For full examples of our prompts, see the appendix.

\begin{figure*}
    \centering
    \includegraphics[width=0.85\linewidth]{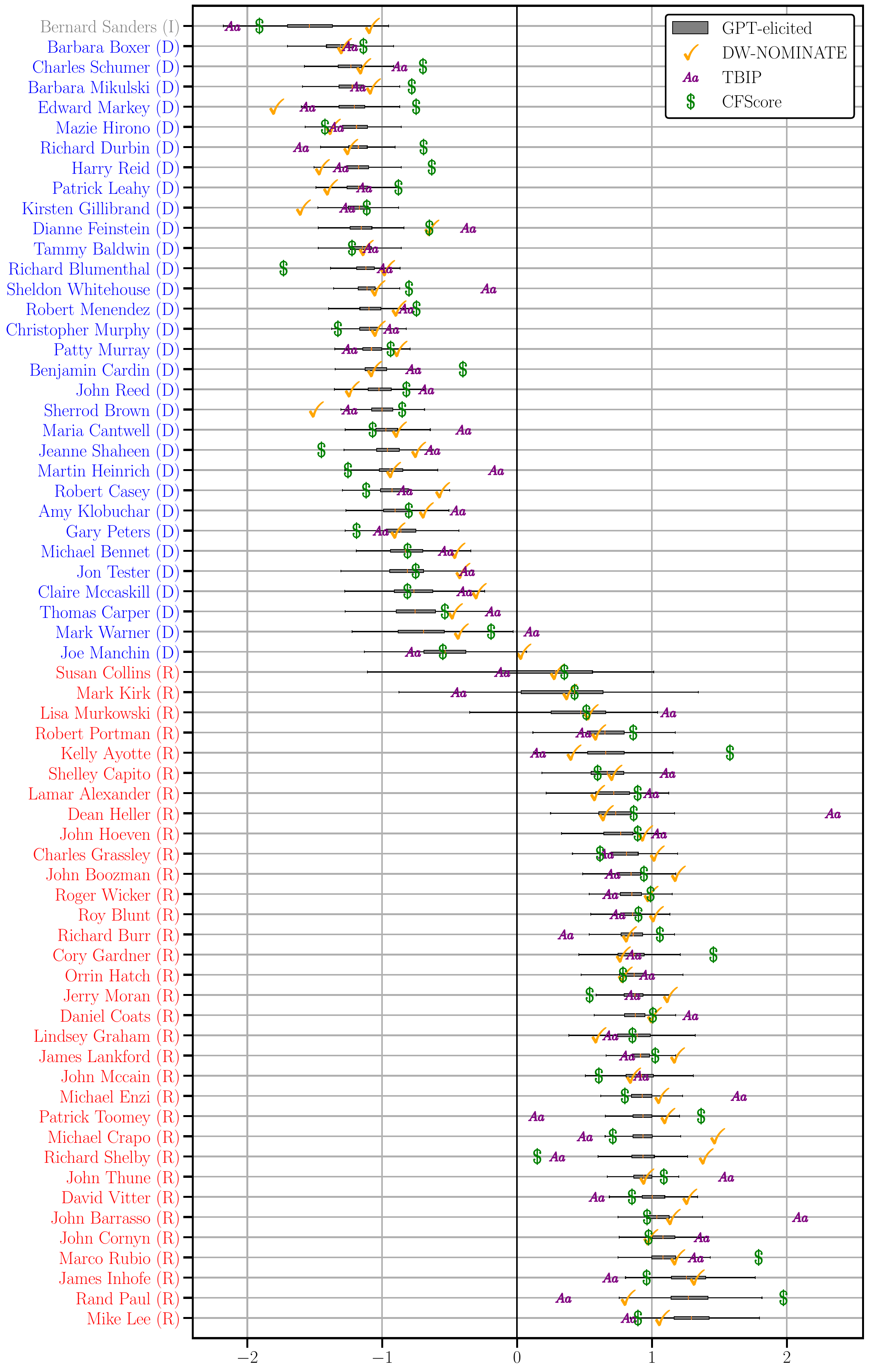}
    \caption{Ideal point scores for members of the $114^{\textrm{th}}$ U.S.~Senate. The GPT-elicited scores are depicted as box-plots, which describe variability across variations in prompt (i.e., Senator order). Rows are sorted in in ascending order by the mean of the GPP-elicited scores. We compare these to those given by DW-NOMINATE \citep{poole_spatial_1985}, TBIP \citep{vafa_text-based_2020}, and CFScores \citep{bonica_mapping_2014}, which exhibit broad correspondence.}
    \label{fig:ideal-point-scores}
\end{figure*}

\begin{figure}
    \centering
    \includegraphics[width=\linewidth]{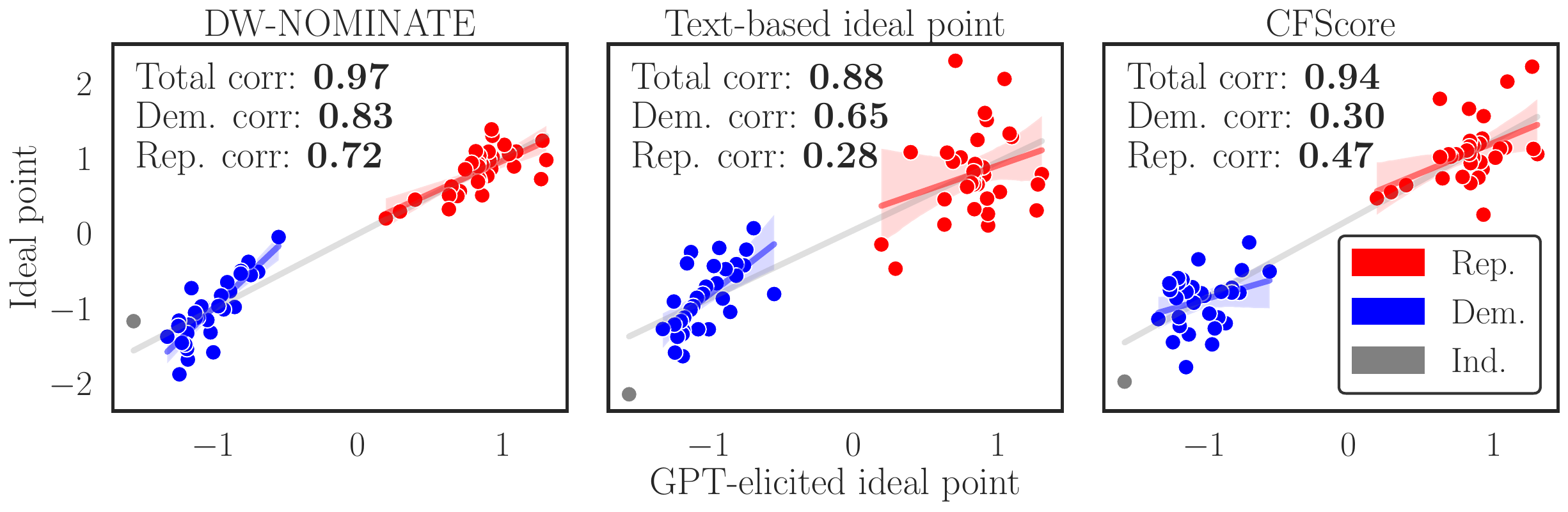}
    \caption{Correlation across and within parties between GPT-elicited ideal points and those of three reference methods.}
    \label{fig:corr}
\end{figure}

\textbf{Face validity.} We plot the ideal points elicited from \texttt{gpt-3.5-turbo} for all Senators in the $114^{\textrm{th}}$ Congress in~\Cref{fig:ideal-point-scores}. Each row contains a box-plot which depicts variability of the elicited ideal point for a given Senator across randomly sampled permutations. We sort the rows in ascending order by the average of these GPT-elicited ideal point and color-code the names of Senators according to their party identification. From this alone we can see that the GPT-elicted ideal points exhibit a high degree of face validity. First, the Democrat and Republican Senators are cleanly separated, with no overlap, and a sharp transition around zero. Second, Bernie Sanders, who is the only Independent and widely recognized to be the most left-leaning Senator, is assigned the most negative ideal point which is substantially left of the second most. Third, the Democrat Joe Manchin and Republican Susan Collins, both of whom are were considered pivotal ``swing'' Senators that sometimes voted with the opposing party, are assigned the two ideal points closest to 0. Many other details beyond these three accord closely with our knowledge.

\textbf{Criterion validity.} As a form of \textit{criterion validity}~\citep{adcock_measurement_2001}, we compared the GPT-elicited ideal points to three well-established models for ideal point estimation. We consider DW-NOMINATE~\citep{poole_d-nominate_2001}, which is based on roll call vote data, CFScores~\citep{bonica_mapping_2014}, which is based on campaign contributions, and text-based ideal points (TBIP)~\citep{vafa_text-based_2020}, which is based on word counts derived from Senators' floor speeches. We regard the ideal points measured by these methods as a form of ``ground truth'', though only loosely---we expect a valid LLM-based measurement approach to not differ too wildly, but still differ in certain meaningful ways, just as we expect these three methods to differ meaningfully among themselves. 

We overlay the ideal points of these reference methods on~\Cref{fig:ideal-point-scores}. Visual inspection suggests broad correspondence between the GPT-elicted ideal points and those of the three references. We analyze this correspondence in~\Cref{fig:corr}, where we give a breakdown of the correlation both within and across Democrats and Republicans. The total correlation of GPT-elicted ideal points with all three is remarkably high---0.97 with DW-NOMINATE, 0.88 with TBIP, and 0.94 CFscores---though this is substantially driven by how all methods easily separate Democrats and Republicans. The within-party correlation with DW-NOMINATE is also high for both parties---0.83 and 0.72---while lower for the other two, with the lowest being TBIP within Republicans. We surmise this relates to~\citet{vafa_text-based_2020}'s finding that word counts were generally less effective for scaling Republicans, who tend to use fewer ideologically-inflected words than Democrats.

\pagebreak
\textbf{Discussion.} It is remarkable to us that an LLM can be used so directly, with simple natural language instruction, to elicit an ideological scaling of legislators that has such face validity and correspondence with other measures. Our findings are consistent with those of~\citet{wu_large_2023} who also find a similar correspondence, and persuasively show that LLMs are not merely ``parroting'' known ideal-point measures. Their approach is more controlled, eliciting only comparisons between pairs of legislators, and then fitting a Bradley-Terry model to infer a scaling implied by the binary outcomes of those comparisons. As such, their approach avoids the issues of scale, anchoring, and sensitivity to ordering that our direct approach must contend with. The trade-off for this reliability is partly scalability, since the pairwise approach depends on querying pairs. The direct approach is also more flexible, allowing for a much broader range measurements to be elicited. We explore this flexibility in the next section, where we deploy direct elicitation to scale the ideology of text in a variety of settings.~\looseness=-1

\section{Eliciting Ideological Scales of Text}
\label{sec:text}

One might choose to be unimpressed that LLMs encode an association between the names of Senators and their ideological leanings. There are undoubtedly many contexts in the training corpus that directly associate politician's names with their political beliefs, and recapitulating such associations might not require a great degree of synthesis. People too are often capable of making vague associations about political leaders---e.g., Bernie Sanders is a socialist---without possessing a sophisticated or well-informed viewpoint.

Ideological slant manifests in language in often subtle and complex ways that require substantial background knowledge and linguistic inference to detect. Recent work in the past two years has begun investigating LLMs' abilities in handling ideological text, like detecting ``dog-whistles''~\citep{mendelsohn_dogwhistles_2023}, rewriting ideological text~\citep{hoyle_making_2023}, and responding coherently to political surveys~\citep{argyle_out_2022}, among others. Most notably, a recent preprint by~\citet{wu_concept-guided_2023} extends the pairwise elicitation approach of~\citet{wu_large_2023} to scale tweets, by eliciting judgements from an LLM whether one of two provided tweets is more liberal. This section contributes to this emerging area by showcasing a variety of ways that a direct elicitation can be flexibly used to scale text and by providing a close-up exploration of the results.

\begin{figure*}[t]
    \centering
    \includegraphics[width=\linewidth]{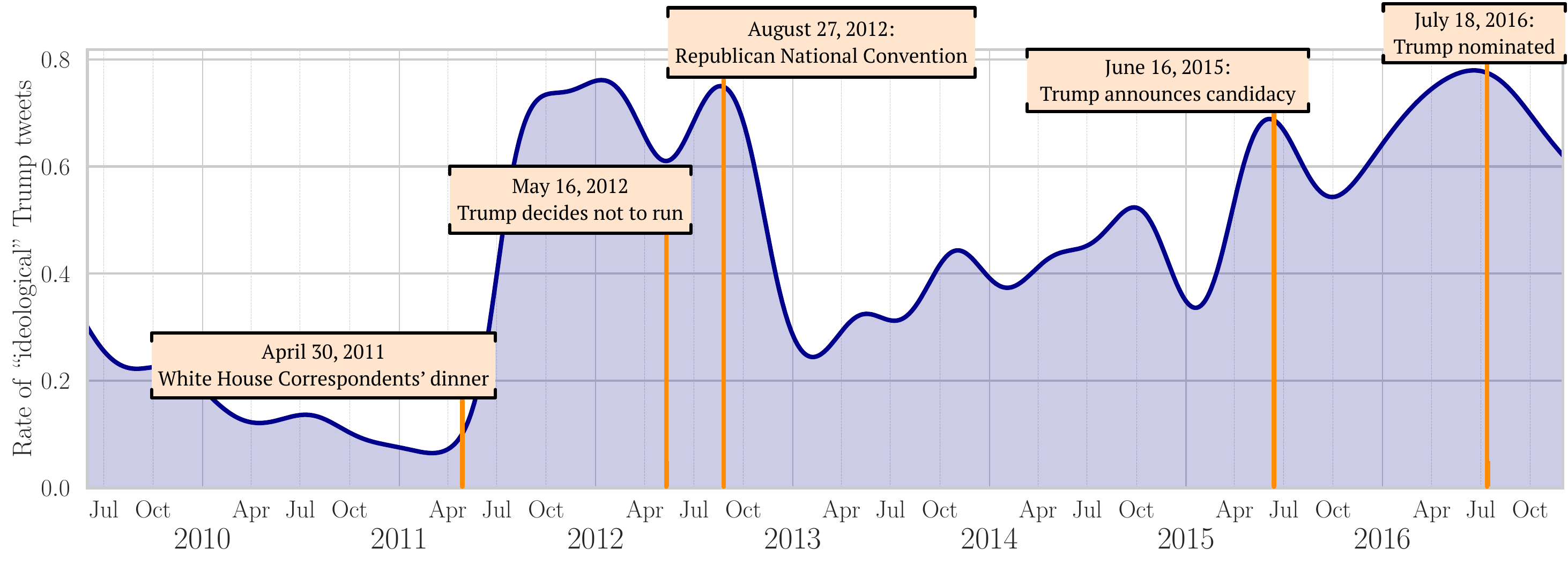}
    \caption{The rate of Trump's tweets being classified by \texttt{gpt-3.5-turbo} as ``ideological'' from 2009 to 2017. A dramatic increase in his rate of ideological tweeting immediately following the 2011 White House Correspondents' dinner is consistent with a long-alleged story about his motivation to run for President in 2016.}
    \label{fig:trumptweets}
\end{figure*}
\subsection{Measuring Trump's political ambitions}
\label{sec:trump}
A natural first question is whether LLMs can reliably distinguish between ideological and non-ideological text. We study this by directly eliciting binary judgements from \texttt{gpt-3.5-turbo}---e.g.
\prompt{
\textbf{Prompt:} Is the following text ideological? Answer yes or no. [...]
}
Our anecdotal experience is that \texttt{gpt-3.5-turbo}'s judgements are remarkably consistent with our own, even in very subtle instances, and we highly encourage the (skeptical) reader to experiment themselves. 

This subsection attempts to validate our impression more systematically, by checking for a form of \textit{construct validity}~\citep{adcock_measurement_2001}. In lieu of ``ground truth'' labels, which would be labor-intensive and time-consuming to produce, we instead check whether patterns in \texttt{gpt-3.5-turbo}'s judgements make sense in relation to other known patterns.

We use a large sample of Donald Trump's tweets from 2009--2017, and ask \texttt{gpt-3.5-turbo} in a few-shot manner whether each one is ideological or not. We then examine whether temporal patterns in Trump's rate of ideological tweeting (as judged by \texttt{gpt-3.5-turbo}) correspond meaningfully to well-known events in Trump's life. The time window begins substantially before 2015, when Trump was not yet in politics (officially), and extends after he took office as President in 2017. It is therefore reasonable to assume that his rate of ideological tweeting changed substantially over that time frame, as he transitioned from business to politics, and that a good method for scaling text will show that.

\Cref{fig:trumptweets} depicts the rate at which \texttt{gpt-3.5-turbo} classified Trump's tweets as ideological over the time frame 2009--2017. To obtain a smooth rate function from the time-stamped binary judgements, we fit a Gaussian process with a squared exponential kernel and logit link function; its mean is visualized. 

The dramatic increase in his rate of ideological tweeting around April 2011 seems to corroborate a widely-alleged story about Trump's decision to run for President. The story, as presented in the PBS documentary ``The Choice 2016''~\citep{frontline_trump}, was that:
\begin{quote}
\begin{quote}
Donald Trump was the focus of President Obama’s jokes at the 2011 White House Correspondents' Dinner. It was there that Trump resolved to run for president [...]
\end{quote}
\end{quote}
This story was also presented in the New Yorker~\citep{new_yorker_trump}, National Review~\citep{natrev_trump}, and TIME~\citep{time_trump}, among several other major media outlets. (We also note that some journalists have cast doubt on the story~\citep{wapo_trump}, and Trump has denied it.) We are unaware of other measurements of Trump's political fixation over time, but this one does appear to show that the time of the dinner was indeed a turning point.  Other aspects of \Cref{fig:trumptweets} also appear to have high validity, like peaks at the times he announced his candidacy and was nominated for President.

\subsection{Detecting a neo-Nazi dog-whistle}
\label{sec:dog}
\begin{figure}[t]
    \centering
    \includegraphics[width=0.75\linewidth]{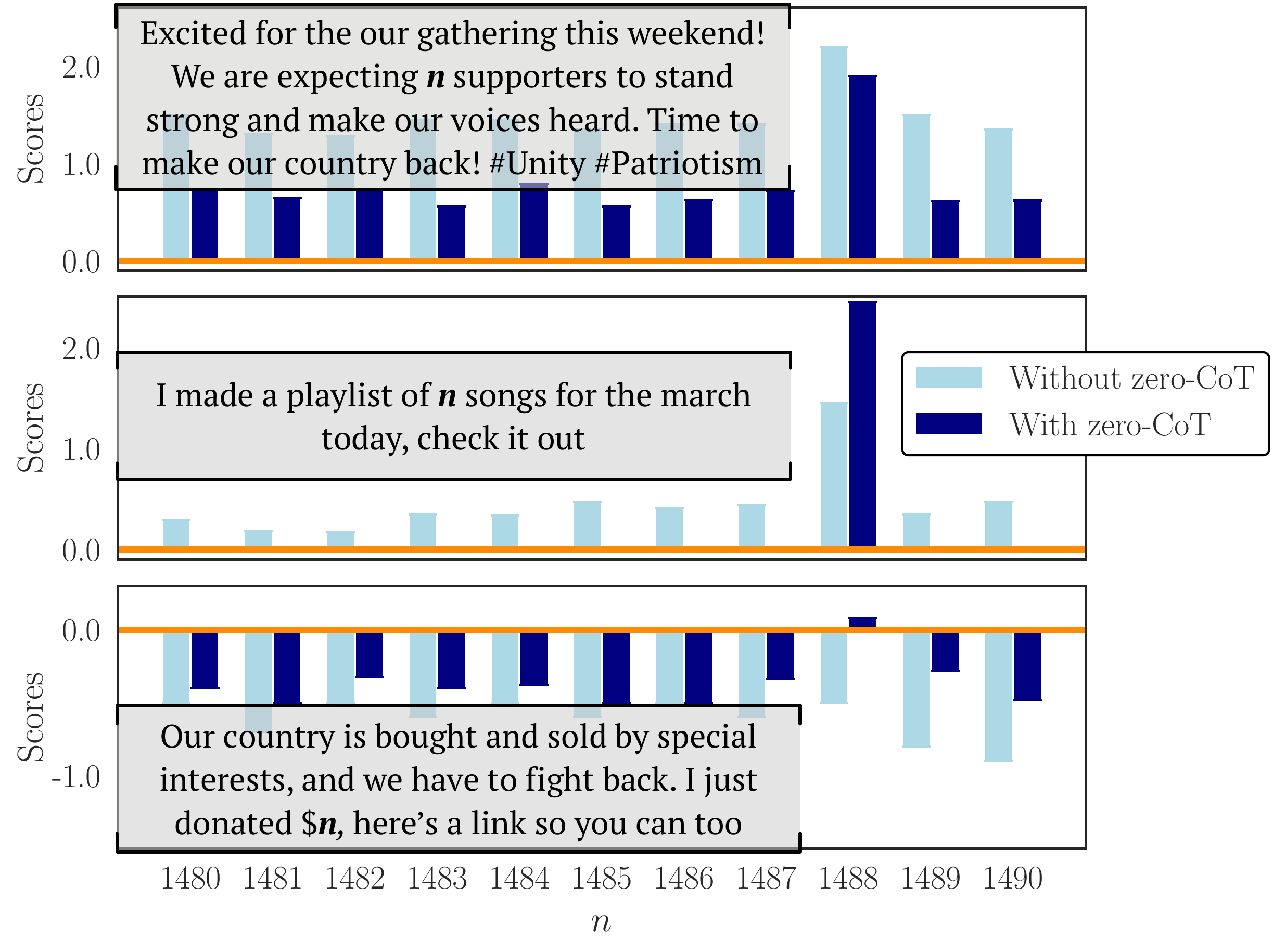}
    \caption{Results on \texttt{gpt-3.5-turbo}'s sensitivity to the dog whistle ``1488'', with and without zero-shot CoT, when scoring the ideology of a vaguely political but otherwise innocuous tweet.}
    \label{fig:dogwhistle}
\end{figure}

Although aggregate trends (and anecdotal experience) suggest that \texttt{gpt-3.5-turbo}'s classification of whether text is ideological has high validity, this leaves open the question of whether it can reliably place ideological text on meaningful scales. This and the next sub-section zoom in on specific hand-crafted examples to build intuition for that question.

Inspired by~\citet{mendelsohn_dogwhistles_2023}, we construct a simple experiment to check if \texttt{gpt-3.5-turbo} picks up on a neo-Nazi ``dog whistle''---i.e., the number ``1488''~\citep{salon_1488}---in otherwise innocuous contexts. We construct three tweet-like examples that are vaguely political. Each is written from the perspective of an organizer who is mobilizing their friends into some form of political activity (e.g., attending a march), and includes reference to some number $n$---e.g.,

\prompt{
    \textbf{Prompt}: Consider the following tweet:\\[0.5em]
    
    I made a playlist of \ul{\textbf{\textit{n}}} songs for the march today, check it out\\[0.5em]
    
    Assign a numerical score which reflects its ideological
    position on a left/right spectrum [...]
}
We try this for $n \in \{1480,\dots,1490\}$, and check whether the scores assigned by \texttt{gpt-3.5-turbo} are substantially different at $n=1488$, which would be consistent with a coded neo-Nazi message.

We visualize these results as light-blue bars in~\Cref{fig:dogwhistle}, where pronounced spikes at $n=1488$ are clearly evident for the first two examples. While the first two are consistent with a right-leaning or moderate interpretation for all numbers, the last is judged by \texttt{gpt-3.5-turbo} to be left-leaning, which we surmise further obscures the neo-Nazi interpretation.

We further experiment by including an encouragement to ``think step-by-step''~\citep{kojima_large_2022}---i.e., 
\prompt{
    \textbf{Prompt}: Please first give your analysis of the tweet and reasoning for the score [...]
}
This kind of prompt engineering is known as ``chain-of-thought'' (CoT) prompting~\citep{wei_chain--thought_2022}, which has been shown to improve performance in a broad array of tasks, including by~\citet{wu_concept-guided_2023} for scaling the ideology of tweets. More specifically, this is a form of \textit{zero-shot CoT}, as the LLM is instructed to provide reasoning and answer within the same response.

The results using zero-shot CoT are visualized in dark-blue in~\Cref{fig:dogwhistle}, where we see that the \texttt{gpt-3.5-turbo} is in fact capable of detecting the dog whistle, even in the harder example. More details on the prompts are provided in the appendix.

\subsection{Dinner table politics}
\label{sec:dinner}
\newcommand{\parent}[1]{{\color{ForestGreen}\ul{\textbf{#1}}}}
\newcommand{\mutter}[1]{{\color{NavyBlue}\ul{\textbf{#1}}}}
\newcommand{\responds}[1]{{\color{BrickRed}\ul{\textbf{#1}}}}

Here we continue with another hand-crafted example that encodes ideology in a much more subtle and diffuse manner. Unlike in the previous example, where knowledge of a single ideologically-inflected word ``1488'' is sufficient, here we construct examples where no individual words or phrases are inherently ideological, but we (the authors) can still form judgements about ideological slant.

In this case, all prompts begin the same by describing a dinner table conversation between Kyle and his parents:

\prompt{
    \textbf{Prompt}: Consider the following scenario:\\[0.5em]
    
    Kyle is home from college, eating dinner with his parents. He mentions that on his walk home, he noticed the new community gardens where people were planting vegetables and herbs. [...]
}
From here the scenario continues on one of 24 possible paths. In each, one of Kyle's two \parent{parents} mutters one of three possible \mutter{exasperated phrases}, and then offers one of four \responds{explanations}. One such path is:
\prompt{
    His \parent{father} rolls \parent{his} eyes and mutters \mutter{``these people''}. Kyle asks his \parent{father} what \parent{he} means. His \parent{father} responds: \responds{``I go to work everyday, and this is what I see.''}
}
From this vignette, we (the authors) feel the father is behaving in a way commonly stereotyped as conservative. We have this feeling despite being unable to point to any specific ideologically-inflected words or other unambiguous indicators. The father's vague appeal to ``work ethic'', a commonly cited conservative value, is perhaps the strongest, but is still rather subtle.~\looseness=-1

Compare this to another possible path:
\prompt{
    His \parent{mother} rolls \parent{her} eyes and mutters \mutter{``these people''}. Kyle asks his \parent{mother} what \parent{she} means. His \parent{mother} responds: \responds{``A bunch of folks used to sleep in that lot, and now they can't because someone wanted to grow heirloom tomatoes.''}
}
From this alternate vignette, we (the authors) interpret the mother to be expressing strong empathy for the homeless, as well as a disdain for the yuppies who displaced them, all of which suggest generally left-leaning political beliefs. Again though, there are no simple words or phrases that are clearly ideologically inflected, and our interpretation requires a combination of linguistic inference and world knowledge.

After describing one of these 24 scenarios, we complete the prompt by instructing the LLM to describe of the parent's political beliefs and assign them an ideal point:

\prompt{
    [...] Use your knowledge as a language model to interpret the \parent{father}’s political beliefs [...] After detailing your reasoning, provide a score that describes \parent{his} beliefs on a Left/Right scale [...]
}

In the appendix, we provide a description of all 24 scenarios. We visualize the scores assigned to each by \texttt{gpt-4} in~\Cref{fig:commgarden}, which accord remarkably closely to our (the authors') impressions. In the appendix, we also include those assigned by \texttt{gpt-3.5-turbo}, which were accordant, but less  so.

\subsection{Scaling the tweets of the $114^{\textrm{th}}$ Senate}
\label{sec:tweets}

In this last case study, we connect back to~\Cref{sec:ideal_points} and investigate whether the scores that \texttt{gpt-3.5-turbo} assigns to a Senator's tweets agrees overall with the ideal point it assigned them based only their name. The absence of such a correspondence would suggest an unsophisticated synthesis of the training corpus, and undermine the validity of both sets of scores.

\textbf{Data set.} The authors~\citet{vafa_text-based_2020} shared with us a data set of tweets of U.S.~legislators from 2009--2017 which is open-source and similar in scope to the one used in their paper on text-based ideal points. For each Senator in the $114^{\textrm{th}}$ Congress, we sub-sampled 250 of their tweets uniformly at random.

\textbf{Basic prompt.} For each tweet in the sub-sampled corpus, we directly elicit an ideological score from \texttt{gpt-3.5-turbo} using prompts of the same flavor as those in~\Cref{sec:dog,sec:dinner}, which ask it to assign a numeric score to the provided tweet using negative numbers for ``left'' and positive for ``right''.~\looseness=-1

\textbf{Dynamic self-anchoring with human feedback.} As with eliciting ideal points (\Cref{sec:ideal_points}), the prompts do not contrain a scale, beyond specifying zero as its midpoint. In the previous section we had the LLM score all Senators at once and relied on its autoregressive nature to ensure that the scores were comparable; we referred to this strategy as self-anchoring. 

\begin{figure}[t]
    \centering
    \includegraphics[width=\linewidth]{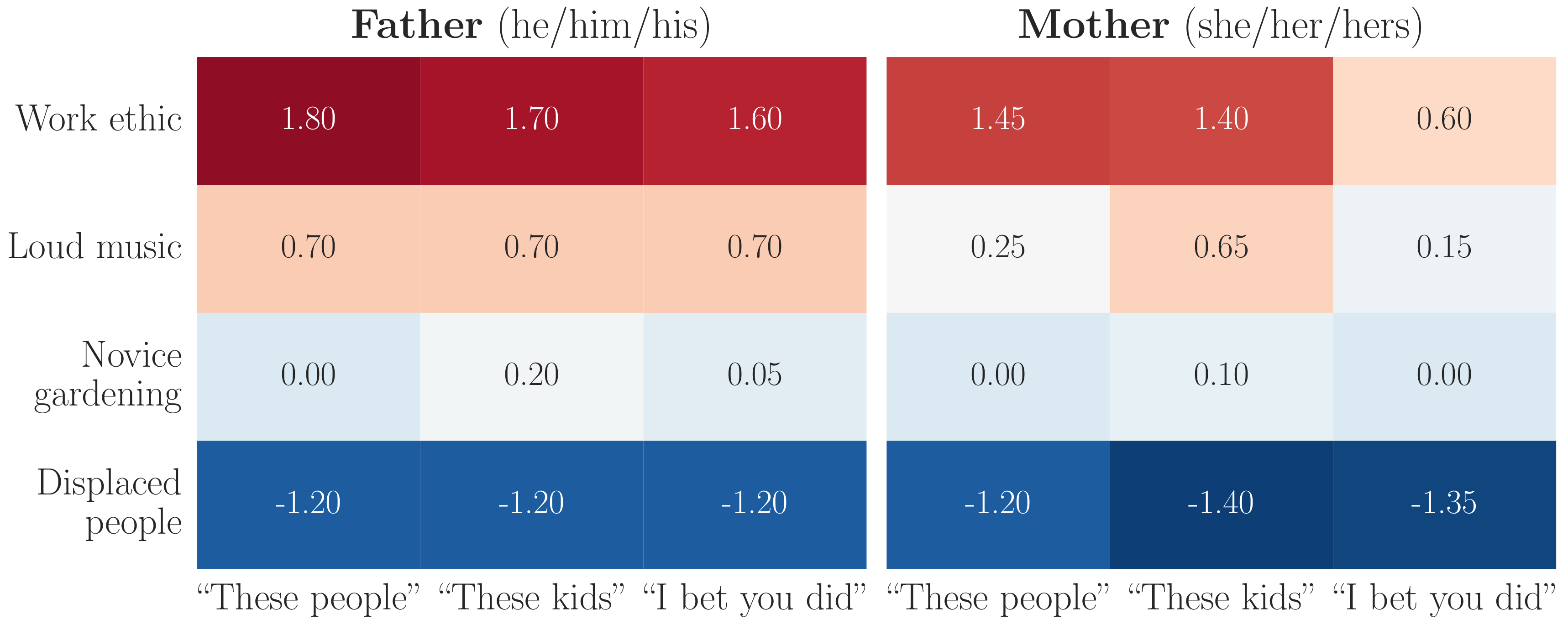}
    \caption{Politics at the dinner table, as elicited from \texttt{gpt-4}. Each subplot corresponds a \parent{parent}, each column an \mutter{exasperated phrases}, and each row an \responds{explanations}, combinations of which subtly alter the interpretation of the parent's ideological slant.}
    \label{fig:commgarden}
\end{figure}

We cannot rely on the same strategy in this instance, since fitting all tweets into a single prompt would greatly exceed \texttt{gpt-3.5-turbo}'s context limit. Instead, for each tweet, we include as context the five previous tweets and scores it assigned to them, a strategy we call dynamic self-anchoring. In addition, we also include four hand-selected tweets and corresponding scores, a form of human feedback. More details are contained in the appendix.

\textbf{Eliciting binary judgements.} We do not expect all text tweeted by a Senator to be intrinsically ideological. We therefore also experiment with first eliciting a binary judgement about whether a tweet is ideological or not (like in~\Cref{sec:trump}), and proceeding to elicit a numeric score only if \texttt{gpt-3.5-turbo}'s response is yes.

\begin{figure*}[t]
    \centering
    \includegraphics[width=\linewidth]{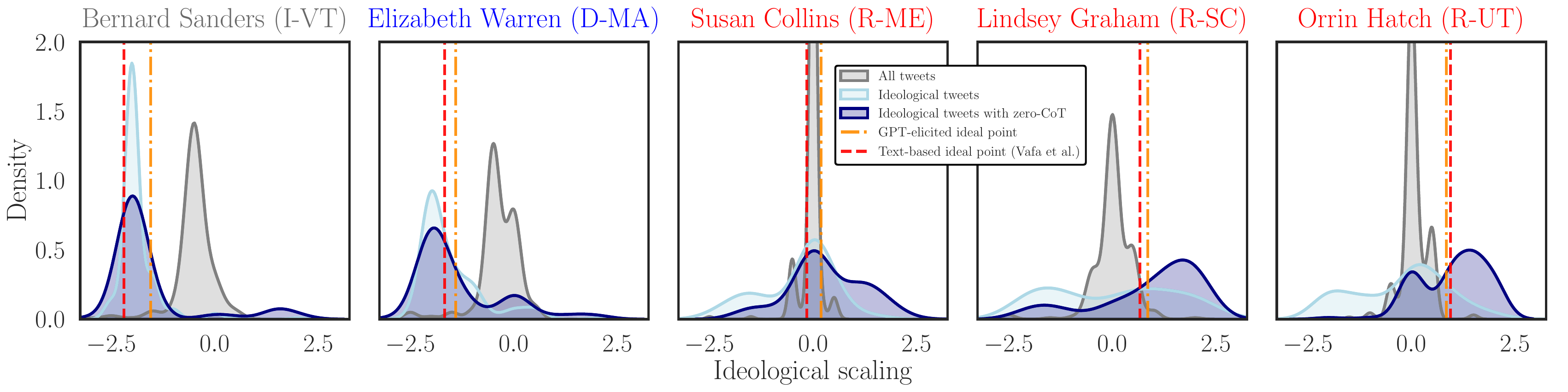}
    \caption{Kernel density estimates of ideal point scalings of sets of tweets by selected senators in the $114^{\textrm{th}}$ U.S. Senate. Tweets are scored using three different prompts: one that scores each tweet, one that scores only tweets that are deemed ideological, and one that scores tweets that are deemed ideological after explaining its reasoning for giving the score.}
    \label{fig:tweetselected}
\end{figure*}

\textbf{Zero-shot chain-of-thought.} As in~\Cref{sec:dinner}, we also experiment with encouraging the LLM to ``think step-by-step'' before providing responses, both to the binary judgements and to the numeric scores. 

\textbf{Overall approaches.} We test three overall approaches for eliciting ideological scores, which all use dynamic self-anchoring with human feedback, but differ in their usage of the other two strategies:~\looseness=-1
\begin{enumerate}
\item ``All tweets'', which does not use any initial binary judgements, or zero-shot CoT in its prompt,
\item ``Ideological tweets'', which introduces initial binary judgements to only score ideological tweets,
\item ``Ideological tweets with zero-CoT'', which adds zero-shot CoT into both elicitation stages.
\end{enumerate}

\begin{figure}
    \centering
    \includegraphics[width=\linewidth]{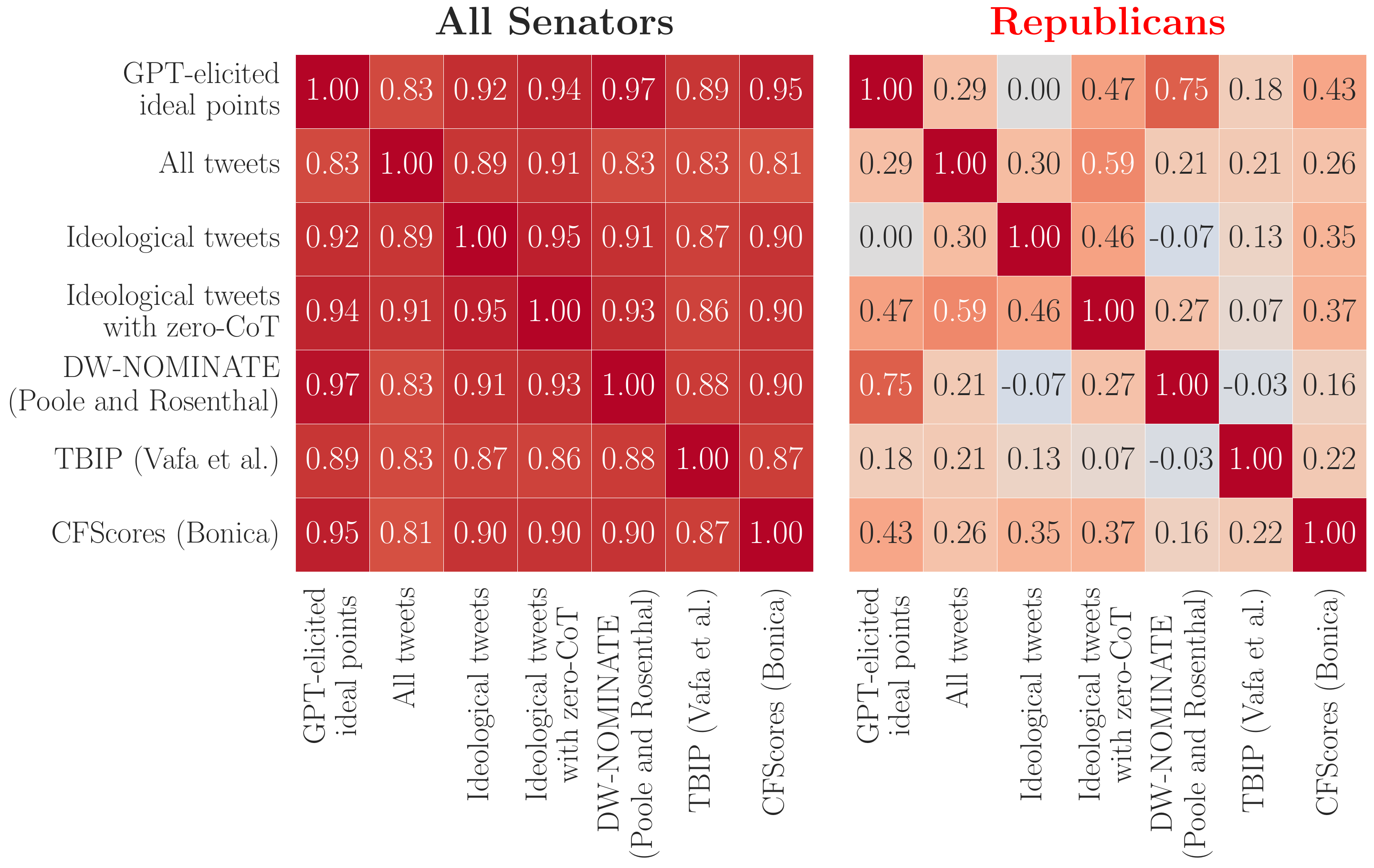}
    \caption{Correlations between average scores of Senators' tweets and GPT-elicited ideal points, as well as with those from the three reference methods. See appendix for within-Democrats results.}
    \label{fig:tweetcorr}
\end{figure}

\textbf{Results.} We first examine the correlation between the average score assigned to all of a Senator's tweets, with the GPT-elicited ideal point assigned to them in \Cref{sec:ideal_points}. We examine this correlation for all three text-scoring methods, and also include the correlations with the three reference methods DW-NOMINATE, TBIP, and CFScores. We further break down correlations across and within parties.

\Cref{fig:tweetcorr} visualizes the overall correlations and those within Republicans (see appendix for those within Democrats). The overall correlations are strikingly high, including 0.94 between the third text-scaling method and the GPT-elicited ideal points, which suggests a remarkable degree of internal consistency. 

The correlations are smaller when broken down by party, which is true between all methods. Intriguingly, among Republicans, the correlations with GPT-elicted ideal points for the three methods are 0.29, 0.0 and 0.47, respectively, which suggests that asking first whether a tweet is ideological or not actually degrades performance if not coupled with zero-shot CoT. This seems consistent with overall picture that ideological slant in Republicans' tweets is diffuse and requires more sophisticated linguistic inference to surface.

To build more intuition, we zoom in on the results for five Senators in~\Cref{fig:tweetselected}, which we found to be generally representative. In each sub-plot, we visualize kernel-density estimates of the distributions of tweet scores by the three different methods. We compare these to the GPT-elicited ideal point and the one assigned by TBIP, which appear as vertical lines. 

From these plots we can clearly see the effect of the initial binary judgements---while the grey densities are all centered at 0, the light blue and dark blue densities are shifted toward the ideal points. We also see that the light blue densities, corresponding to the second method without zero-shot CoT, are peaked around the ideal points for the three left-most Senators, but not for the two right-most. Meanwhile the dark blue densities, which correspond to the third method with zero-shot CoT, appear to concentrate in the area of the ideal points for all Senators. These results seem to corroborate the aggregate correlations, that the most sophisticated tweet-scaling method has a high degree of internal consistency with the GPT-elicited ideal points.

\pagebreak
\section{Generating and scaling policy platforms}
\label{sec:generate}
In this last section, we report experiments in much more open-ended setting involving \textit{text generation}. In these experiments, we ask models to generate text describing the policy positions that a lawmaker with a given ideal point (e.g., +3) might support on a given issue (e.g., climate and environment). We then ask another model to score the given policy position on that issue, thus further examining how internally consistent models' judgements about ideology are, even when generating open-ended text.~\looseness=-1

\textbf{Generating platforms.} For a set of 18 issues enumerated on the U.S. Department of State website (e.g., energy, climate crisis, human trafficking), we prompt \texttt{gpt-4-turbo} to generate a detailed description of policy positions that a lawmaker with a given ideal point (e.g., +3) might support. To anchor the scale, we first instruct the model to provide ideal points for the Senators in the $114^\textrm{th}$ Congress using the same methodology described in~\Cref{sec:ideal_points}. With that pairing of Senators and ideal points then in its context, we instruct it to generate a policy platform for a given issue at a score randomly selected from its elicited ideal points---an example of this procedure is given below:
\prompt{
\textbf{Prompt:} Assign a numerical ideal point score to each member of the following list of
United States politicians [...]

\textbf{GPT-4:} Donald Trump: 7.5 [...] Joe Donnelly: -2.5 [...] Elizabeth Warren: -7.5 [...]

\textbf{Prompt:} Using these scores as your scale, give an example of a policy position on the topic of \textbf{economic prosperity and trade policy} that someone with a score of \textbf{-2.5} might have [...]

\textbf{GPT-4:} 
[...] Support the negotiation of trade agreements that open up foreign markets to domestic products, thereby promoting job growth at home. However, these agreements must include enforceable labor standards and environmental protections to prevent a race to the bottom and ensure that [...]

}
\textbf{Anchoring.} We do not actually have every model assign ideal points to the $114^{\textrm{th}}$ Senate every time. Rather, we elicit the ideal points once, and then for each subsequent model, we enter the prompt and response into its context, so that it proceeds as if it had itself assigned those exact ideal points to Senators. In so doing, we anchor the scale both within and across policy platform generations.

\textbf{Scaling platforms.} For each model-generated policy platform on each issue, we then prompt another model to score the platform on an ideological scale, again registering in its context the same assignment of ideal points to the $114^{\textrm{th}}$ Senate---e.g.,
\prompt{
\textbf{Prompt:} Assign a numerical ideal point score to each member of the following list of
United States politicians [...]

\textbf{GPT-4:} Donald Trump: 7.5 [...] Joe Donnelly: -2.5 [...] Elizabeth Warren: -7.5 [...]

\textbf{Prompt:} Using the previous scores as your scale, provide an ideological left-right score to someone who holds the following policy position on the topic of \textbf{economic prosperity and trade policy}: [...]

\textbf{GPT-4:} -3.0
}
By anchoring all models to the same numerical ideological scale, we can then compare the scores at which one model generated a policy position to the scores another model rated them to investigate how consistent a models' ideological judgements are with its own open-ended text generations.

\begin{figure}
    \centering
    \includegraphics[width=\textwidth]{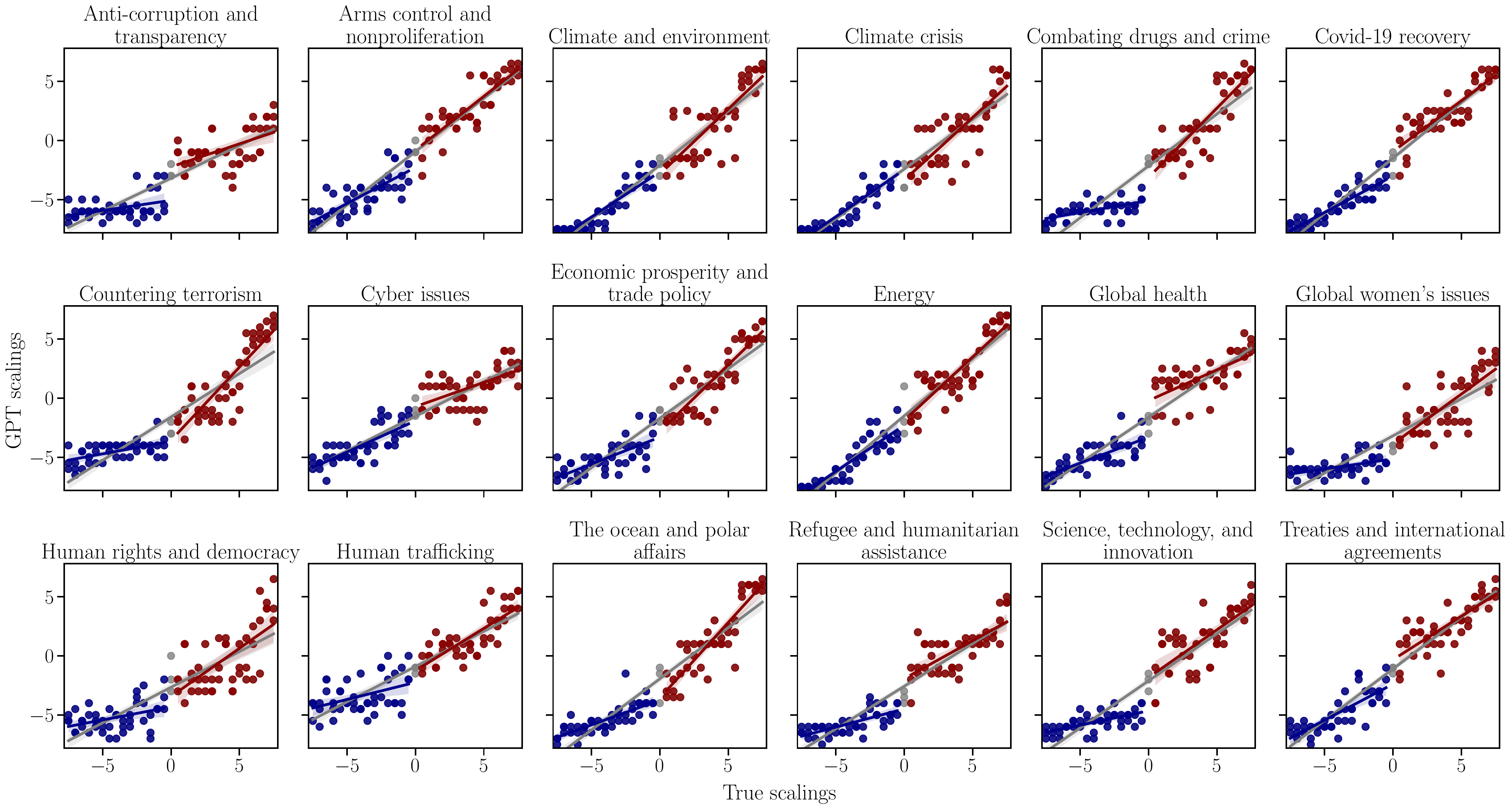}
    \caption{Scalings assigned by \texttt{gpt-4-turbo} to ideological positions compared with the true scaling from which the position was generated, separated by issue. Regression lines are shown both for within (red, blue) and across (grey) party. We see a high correlation across party and within Republicans, while for Democrats there is lower correlation (though always positive) on certain issues.}
    \label{fig:scalings}
\end{figure}

\textbf{Zero-shot chain-of-thought.} As in~\Cref{sec:dinner} and ~\Cref{sec:tweets}, we prompt the LLM to ``think step-by-step'' before providing responses, both when writing a policy position for a given numerical value, and when assigning a numerical value back to a given policy position.

\textbf{Results.} We visualize the results in~\Cref{fig:scalings}. Overall, we observe a very high correspondence between the ``true'' scores used to generate the policies and the scores later assigned to the policies. More specifically, for all issues, we observe a positive slope in the regression lines, many of which are close to the diagonal, indicating positive correlation between the two sets of scores.~\looseness=-1

We measure correlation both across parties (grey) and within Democrats (blue) and Republicans (red). The across-party correlations are uniformly high across all issues, suggesting that the model is highly internally consistent in discriminating broadly ``left'' versus ``right'' policy positions. We furthermore observe that the within-Republican correlations are also high across all issues, suggesting internal consistency in discriminating different levels of ``right'' policy positions. 

While the within-Democrat correlations are always positive, there are a handful of issues for which they are small, like \textit{anti-corruption and transparency} (top left-most) or \textit{global women's issues} (center right-most). We should not necessarily expect that all issues have a spectrum of well-differentiated policy positions within parties, and wonder whether these are issues for which Democrats' exhibit little differentiation in their policy preferences. We leave a full investigation of this phenomenon to future work, but include below in~\cref{fig:twocolumns} an example of two generated platforms for \textit{combating drugs and crime}, both supposedly ``left-wing'', which the scoring model was unable to distinguish well. For reference, we also provide an example in~\cref{fig:twocolumns-left-right} of a ``left'' and ``right'' policy platform, that exhibit a degree of face validity, and which the scoring model was able to distinguish.

\begin{figure}[t!]
\centering
% First Column
\begin{minipage}[t]{0.4\textwidth}
\centering
\fbox{ 
\begin{minipage}{\textwidth}
\textbf{Position 1}
\par\medskip
\textit{Implement comprehensive community-based programs that prioritize prevention and treatment over incarceration for non-violent drug offenses. Support initiatives that provide education, job training, and mental health services to at-risk populations to reduce the likelihood of drug abuse and criminal activity. Advocate for the decriminalization of certain drugs and the expungement of records for individuals with past non-violent drug offenses to facilitate their reintegration into society.}~\looseness=-1

\end{minipage}
}
\end{minipage}
\hfill % Spacing between the minipages
% Second Column
\begin{minipage}[t!]{0.52\textwidth}
\centering
\fbox{
\begin{minipage}{\textwidth}
\textbf{Position 2}
\par\medskip 
\textit{Implement a comprehensive approach to drug offenses that prioritizes treatment and rehabilitation over incarceration. Decriminalize possession of small amounts of drugs for personal use, and invest in substance abuse treatment programs that are accessible and affordable for all. Redirect funds from the enforcement of drug prohibition toward community-based initiatives that address the socioeconomic factors contributing to drug abuse and crime. Reform the criminal justice system to eliminate mandatory minimum sentences for non-violent drug offenses, ensure fair sentencing practices, and invest in restorative justice programs that focus on the reintegration of offenders into society. Establish oversight mechanisms to reduce racial disparities in arrests, sentencing, and incarceration.}
\end{minipage}
}
\end{minipage}

\caption{Two policy positions generated by \texttt{gpt-4-turbo} on the issue of \textit{combating drugs and crime}. Position 1 was generated for a score of -1.0, and position 2 for a score of -6.0 (both ``left-wing''). Both positions were later scored at -6.0. This example highlights the small relative difference in these positions. Both advocate for a shift from punitive to rehabilitative approaches to non-violent drug offenses, emphasizing treatment, education, and reintegration over incarceration.}
\label{fig:twocolumns}
\end{figure}

\begin{figure}[t!]
\centering
% First Column
\begin{minipage}[t]{0.48\textwidth}
\centering
\fbox{ 
\begin{minipage}{\textwidth}
\textbf{Position 1}
\par\medskip
       \textit{Implement a comprehensive climate action plan that \textbf{aggressively reduces greenhouse gas emissions} through a combination of regulations and incentives. This includes transitioning to  renewable energy sources by: investing in energy efficiency across all sectors, and supporting research and development in clean technology. Establish a carbon pricing mechanism to reflect the true cost of carbon emissions, and use the revenue to fund renewable energy projects and assist communities and workers \textbf{transitioning from fossil fuel} industries. Rejoin and actively participate in international climate agreements to ensure a global commitment to reducing emissions and combating climate change.}
~\looseness=-1

\end{minipage}
}
\end{minipage}
\hfill % Spacing between the minipages
% Second Column
\begin{minipage}[t]{0.46\textwidth}
\centering
\fbox{
\begin{minipage}{\textwidth}
\textbf{Position 2}
\par\medskip 
              \textit{Support the \textbf{gradual transition to cleaner energy} sources through market incentives and innovation. Encourage private investment in renewable energy technologies and provide tax credits for companies that reduce their carbon footprint. However, \textbf{maintain a strong commitment to domestic fossil fuel} production to ensure energy independence and economic security. Oppose any drastic measures that would lead to significant job losses or increase energy costs for consumers.}
\end{minipage}
}
\end{minipage}

\caption{Two policy positions generated by \texttt{gpt-4-turbo} on the issue of \textit{climate crisis}. Position 1 was generated for a score of -4.5 (``left-wing''), and position 2 for a score of 4.5 (``right-wing''). Position 1 was assigned with a score of -4.5 and later given a score of -7.0. Position 2 was generated for a score of 4.5, and later scored at 2.0. On their face, both positions seem clearly identifiable as ``left'' and ``'right'', based on the bolded excerpts, among other details.}
\label{fig:twocolumns-left-right}
\end{figure}

\section{Conclusion}
\label{sec:discussion}
Many share the general perspective that LLMs will reshape the social and political sciences continues~\citep[among others]{bail2023can, grossmann2023ai,linegar_large_2023,ziems_can_2023}.
Our perspective is that LLMs specifically prompt a fundamental rethinking of measurement in the social sciences, and more specifically, the role of quantification in measurement. LLMs allow social scientists to employ and refer to complex constructs in a flexible and scalable manner, all within natural language. A social scientist who wishes to use terms like ``left'' and ``right''-wing loosely, as one might with another human, can now do so with machines, which do not require the scientist to commit themself to overly specific formal definitions or mathematical operationalizations. LLMs exhibit remarkable fluency in navigating underspecification, an inherent property of natural language~\citep{ludlow2014living}. Our view is that this opens a new mode of social scientific research, that is not well described as quantitative or qualitative, whereby qualitative-like judgements and analyses are made in partnership with a cooperative machine that can scale them to massive data sets, and repeat them across a myriad of controlled conditions. The purpose of this paper is not to propose a particular method, but rather to provide a suite of illustrative examples of how LLMs can be flexibly used in this manner, and to show evidence of their promise.

There are challenges and perils. LLMs' qualitative judgements should never be simply trusted, or regarded as ``objective''. Rather, their responses should be continuously scrutinized, validated, and checked for inconsistency. This paper attempts to demonstrate how a social scientist might employ a variety of tests to check for face validity, convergent validity, criterion and construct validity, and internal consistency, among other properties that lend credibility to  measurements~\citep{adcock_measurement_2001,jacobs_measurement_2021}.

An important question we have thus far left unaddressed is: when eliciting ideological scales (or any other kind of social scientific scale), what exactly is our estimand~\citep{lundberg2021your}? Traditional quantitative methods for scaling ideology operate solely on objective data of lawmakers' behavior---such as their votes, contributor history, or tweets---and thus seek to estimate an objective quality of the lawmaker. The methodology employed in this paper cannot be credibly understood as targeting such estimands since, by contrast, LLMs' training corpora contain a devilish mix of punditry, opinion, a mix of true and false assertions about the lawmakers, and so on. Leaving aside the question of whether they ever could be used to target such estimands, we view LLMs instead simply as ``Zeitgeist machines'', which synthesize not what is true (e.g., of a given lawmaker) but rather what the public believes is true. In this regard, we resonate with the findings of~\citet{wu_large_2023}, who find that their ``ChatScores'' correlate most closely with the ``Perceived Ideology scores'' of~\citet{hopkins2022trump}, which measure ``voters’ or activists’ \textit{perceptions} of [lawmakers' ideological positions]''. Whether LLMs can be used to synthesize public opinion raises the question of ``which public''~\citep{santurkar_whose_2023}, among others. There many challenges ahead in using LLMs to reshape measurement---we hope this paper serves as a call to face them.

\section*{Acknowledgment}
We thank (in alphabetical order) Alexander Hoyle, Shun Kamaya, Chris Kennedy, Barry Schein, Brandon Stewart, Niklas Stoehr, Keyon Vafa, and Victor Veitch for discussions and feedback.

\section*{References}
\renewcommand{\refname}{}
\renewcommand{\refname}{\vspace{-\baselineskip}}
\renewcommand{\bibname}{\vspace{-\baselineskip}}
\bibliographystyle{plainnat}
\bibliography{zotero,references}

\begin{thebibliography}{59}
\providecommand{\natexlab}[1]{#1}
\providecommand{\url}[1]{\texttt{#1}}
\expandafter\ifx\csname urlstyle\endcsname\relax
  \providecommand{\doi}[1]{doi: #1}\else
  \providecommand{\doi}{doi: \begingroup \urlstyle{rm}\Url}\fi

\bibitem[Adcock and Collier(2001)]{adcock_measurement_2001}
Robert Adcock and David Collier.
\newblock Measurement {Validity}: {A} {Shared} {Standard} for {Qualitative} and {Quantitative} {Research}.
\newblock \emph{American Political Science Review}, 95\penalty0 (3):\penalty0 529--546, September 2001.

\bibitem[Andreas(2022)]{andreas_language_2022}
Jacob Andreas.
\newblock Language {Models} as {Agent} {Models}, December 2022.
\newblock URL \url{http://arxiv.org/abs/2212.01681}.
\newblock arXiv:2212.01681 [cs].

\bibitem[Argyle et~al.(2022)Argyle, Busby, Fulda, Gubler, Rytting, and Wingate]{argyle_out_2022}
Lisa~P. Argyle, Ethan~C. Busby, Nancy Fulda, Joshua Gubler, Christopher Rytting, and David Wingate.
\newblock Out of {One}, {Many}: {Using} {Language} {Models} to {Simulate} {Human} {Samples}.
\newblock In \emph{Proceedings of the 60th {Annual} {Meeting} of the {Association} for {Computational} {Linguistics} ({Volume} 1: {Long} {Papers})}, pages 819--862, 2022.
\newblock \doi{10.18653/v1/2022.acl-long.60}.
\newblock URL \url{http://arxiv.org/abs/2209.06899}.
\newblock arXiv:2209.06899 [cs].

\bibitem[Bafumi et~al.(2005)Bafumi, Gelman, Park, and Kaplan]{bafumi_practical_2005}
Joseph Bafumi, Andrew Gelman, David~K. Park, and Noah Kaplan.
\newblock Practical {Issues} in {Implementing} and {Understanding} {Bayesian} {Ideal} {Point} {Estimation}.
\newblock \emph{Political Analysis}, 13\penalty0 (2):\penalty0 171--187, 2005.
\newblock ISSN 1047-1987, 1476-4989.
\newblock \doi{10.1093/pan/mpi010}.
\newblock URL \url{https://www.cambridge.org/core/product/identifier/S1047198700001054/type/journal_article}.

\bibitem[Bail(2023)]{bail2023can}
Christopher~A Bail.
\newblock Can generative ai improve social science?
\newblock 2023.

\bibitem[Barberá(2015)]{barbera_birds_2015}
Pablo Barberá.
\newblock Birds of the {Same} {Feather} {Tweet} {Together}: {Bayesian} {Ideal} {Point} {Estimation} {Using} {Twitter} {Data}.
\newblock \emph{Political Analysis}, 23\penalty0 (1):\penalty0 76--91, 2015.
\newblock ISSN 1047-1987, 1476-4989.
\newblock \doi{10.1093/pan/mpu011}.
\newblock URL \url{https://www.cambridge.org/core/product/identifier/S104719870001161X/type/journal_article}.

\bibitem[Bender et~al.(2021)Bender, Gebru, McMillan-Major, and Shmitchell]{bender2021dangers}
Emily~M Bender, Timnit Gebru, Angelina McMillan-Major, and Shmargaret Shmitchell.
\newblock On the dangers of stochastic parrots: Can language models be too big?
\newblock In \emph{Proceedings of the 2021 ACM conference on fairness, accountability, and transparency}, pages 610--623, 2021.

\bibitem[Bonica(2014)]{bonica_mapping_2014}
Adam Bonica.
\newblock Mapping the {Ideological} {Marketplace}.
\newblock \emph{American Journal of Political Science}, 58\penalty0 (2), April 2014.

\bibitem[Bonica(2019)]{bonica_are_2019}
Adam Bonica.
\newblock Are {Donation}-{Based} {Measures} of {Ideology} {Valid} {Predictors} of {Individual}-{Level} {Policy} {Preferences}?
\newblock \emph{The Journal of Politics}, 81\penalty0 (1):\penalty0 327--333, January 2019.
\newblock ISSN 0022-3816, 1468-2508.
\newblock \doi{10.1086/700722}.
\newblock URL \url{https://www.journals.uchicago.edu/doi/10.1086/700722}.

\bibitem[Clinton et~al.(2004)Clinton, Jackman, and Rivers]{clinton_statistical_2004}
Joshua Clinton, Simon Jackman, and Douglas Rivers.
\newblock The {Statistical} {Analysis} of {Roll} {Call} {Data}.
\newblock \emph{American Political Science Review}, 98\penalty0 (2):\penalty0 355--370, May 2004.
\newblock ISSN 0003-0554, 1537-5943.
\newblock \doi{10.1017/S0003055404001194}.
\newblock URL \url{https://www.cambridge.org/core/product/identifier/S0003055404001194/type/journal_article}.

\bibitem[Converse(2006)]{converse_nature_2006}
Philip~E. Converse.
\newblock The nature of belief systems in mass publics (1964).
\newblock \emph{Critical Review}, 18\penalty0 (1-3):\penalty0 1--74, January 2006.
\newblock ISSN 0891-3811, 1933-8007.
\newblock \doi{10.1080/08913810608443650}.
\newblock URL \url{http://www.tandfonline.com/doi/abs/10.1080/08913810608443650}.

\bibitem[Frontline(2016)]{frontline_trump}
PBS Frontline.
\newblock The choice 2016.
\newblock Frontline, 2016.
\newblock URL \url{https://www.pbs.org/wgbh/frontline/documentary/the-choice-2016/?}
\newblock PBS.

\bibitem[Gajanan(2017)]{time_trump}
Mahita Gajanan.
\newblock A history of president trump being trolled at the white house correspondents' dinner.
\newblock \emph{TIME}, April 2017.
\newblock URL \url{https://time.com/4756751/donald-trump-white-house-correspondents-dinner/}.

\bibitem[Gerring(1997)]{ideologyGerring1997}
John Gerring.
\newblock Ideology: {A} {Definitional} {Analysis}.
\newblock \emph{Political Research Quarterly}, 50\penalty0 (4):\penalty0 957--994, December 1997.

\bibitem[Gerrish and Blei(2012)]{gerrish_issue-adjusted_2012}
Sean~M. Gerrish and David~M. Blei.
\newblock The {Issue}-{Adjusted} {Ideal} {Point} {Model}, September 2012.
\newblock URL \url{http://arxiv.org/abs/1209.6004}.
\newblock arXiv:1209.6004 [cs, stat].

\bibitem[Gopnik(2015)]{new_yorker_trump}
Adam Gopnik.
\newblock A history of president trump being trolled at the white house correspondents' dinner.
\newblock \emph{The New Yorker}, September 2015.
\newblock URL \url{https://www.newyorker.com/news/daily-comment/trump-and-obama-a-night-to-remember}.

\bibitem[Grimmer and Stewart(2013)]{grimmer2013text}
Justin Grimmer and Brandon~M Stewart.
\newblock Text as data: The promise and pitfalls of automatic content analysis methods for political texts.
\newblock \emph{Political analysis}, 21\penalty0 (3):\penalty0 267--297, 2013.

\bibitem[Grossmann et~al.(2023)Grossmann, Feinberg, Parker, Christakis, Tetlock, and Cunningham]{grossmann2023ai}
Igor Grossmann, Matthew Feinberg, Dawn~C Parker, Nicholas~A Christakis, Philip~E Tetlock, and William~A Cunningham.
\newblock Ai and the transformation of social science research.
\newblock \emph{Science}, 380\penalty0 (6650):\penalty0 1108--1109, 2023.

\bibitem[Gu et~al.(2014)Gu, Sun, Jiang, Wang, and Chen]{gu_topic-factorized_2014}
Yupeng Gu, Yizhou Sun, Ning Jiang, Bingyu Wang, and Ting Chen.
\newblock Topic-factorized ideal point estimation model for legislative voting network.
\newblock In \emph{Proceedings of the 20th {ACM} {SIGKDD} international conference on {Knowledge} discovery and data mining}, pages 183--192, New York New York USA, August 2014. ACM.
\newblock ISBN 978-1-4503-2956-9.
\newblock \doi{10.1145/2623330.2623700}.
\newblock URL \url{https://dl.acm.org/doi/10.1145/2623330.2623700}.

\bibitem[Hand(1996)]{hand_statistics_1996}
D.~J. Hand.
\newblock Statistics and the {Theory} of {Measurement}.
\newblock \emph{Journal of the Royal Statistical Society. Series A (Statistics in Society)}, 159\penalty0 (3):\penalty0 445, 1996.
\newblock ISSN 09641998.
\newblock \doi{10.2307/2983326}.
\newblock URL \url{https://www.jstor.org/stable/10.2307/2983326?origin=crossref}.

\bibitem[Hopkins and Noel(2022)]{hopkins2022trump}
Daniel~J Hopkins and Hans Noel.
\newblock Trump and the shifting meaning of “conservative”: Using activists’ pairwise comparisons to measure politicians’ perceived ideologies.
\newblock \emph{American Political Science Review}, 116\penalty0 (3):\penalty0 1133--1140, 2022.

\bibitem[Hoyle et~al.(2023)Hoyle, Sarkar, Goel, and Resnik]{hoyle_making_2023}
Alexander Hoyle, Rupak Sarkar, Pranav Goel, and Philip Resnik.
\newblock Making the {Implicit} {Explicit}: {Implicit} {Content} as a {First} {Class} {Citizen} in {NLP}, May 2023.
\newblock URL \url{http://arxiv.org/abs/2305.14583}.
\newblock arXiv:2305.14583 [cs].

\bibitem[Jackman(2001)]{jackman_multidimensional_2001}
Simon Jackman.
\newblock Multidimensional {Analysis} of {Roll} {Call} {Data} via {Bayesian} {Simulation}: {Identification}, {Estimation}, {Inference}, and {Model} {Checking}.
\newblock \emph{Political Analysis}, 9\penalty0 (3):\penalty0 227--241, January 2001.
\newblock ISSN 1047-1987, 1476-4989.
\newblock \doi{10.1093/polana/9.3.227}.
\newblock URL \url{https://www.cambridge.org/core/product/identifier/S1047198700003818/type/journal_article}.

\bibitem[Jacobs and Wallach(2021)]{jacobs_measurement_2021}
Abigail~Z. Jacobs and Hanna Wallach.
\newblock Measurement and {Fairness}.
\newblock In \emph{Proceedings of the 2021 {ACM} {Conference} on {Fairness}, {Accountability}, and {Transparency}}, pages 375--385, March 2021.
\newblock \doi{10.1145/3442188.3445901}.
\newblock URL \url{http://arxiv.org/abs/1912.05511}.
\newblock arXiv:1912.05511 [cs].

\bibitem[Kinder and Kalmoe(2017)]{kinder_neither_2017}
Donald~R. Kinder and Nathan~P. Kalmoe.
\newblock \emph{Neither {Liberal} {Nor} {Conservative}: {Ideological} {Innocence} in the {American} {Public}}.
\newblock Chicago studies in {American} politics. The University of Chicago Press, Chicago ; London, 2017.
\newblock ISBN 978-0-226-45231-9 978-0-226-45245-6.

\bibitem[Kojima et~al.(2022)Kojima, Gu, Reid, Matsuo, and Iwasawa]{kojima_large_2022}
Takeshi Kojima, Shixiang~Shane Gu, Machel Reid, Yutaka Matsuo, and Yusuke Iwasawa.
\newblock Large {Language} {Models} are {Zero}-{Shot} {Reasoners}.
\newblock 2022.

\bibitem[Krippendorff(2018)]{krippendorff2018content}
Klaus Krippendorff.
\newblock \emph{Content analysis: An introduction to its methodology}.
\newblock Sage publications, 2018.

\bibitem[Kuhn(1962)]{kuhn1962structure}
Thomas~S Kuhn.
\newblock The structure of scientific revolutions.
\newblock \emph{The University of Chicago Press}, 2:\penalty0 90, 1962.

\bibitem[Lauderdale and Clark(2014)]{lauderdale_scaling_2014}
Benjamin~E. Lauderdale and Tom~S. Clark.
\newblock Scaling {Politically} {Meaningful} {Dimensions} {Using} {Texts} and {Votes}.
\newblock \emph{American Journal of Political Science}, 58\penalty0 (3):\penalty0 754--771, July 2014.
\newblock ISSN 00925853.
\newblock \doi{10.1111/ajps.12085}.
\newblock URL \url{https://onlinelibrary.wiley.com/doi/10.1111/ajps.12085}.

\bibitem[Linegar et~al.(2023)Linegar, Kocielnik, and Alvarez]{linegar_large_2023}
Mitchell Linegar, Rafal Kocielnik, and R.~Michael Alvarez.
\newblock Large language models and political science.
\newblock \emph{Frontiers in Political Science}, 5:\penalty0 1257092, October 2023.
\newblock ISSN 2673-3145.
\newblock \doi{10.3389/fpos.2023.1257092}.
\newblock URL \url{https://www.frontiersin.org/articles/10.3389/fpos.2023.1257092/full}.

\bibitem[Ludlow(2014)]{ludlow2014living}
Peter Ludlow.
\newblock \emph{Living words: Meaning underdetermination and the dynamic lexicon}.
\newblock OUP Oxford, 2014.

\bibitem[Lundberg et~al.(2021)Lundberg, Johnson, and Stewart]{lundberg2021your}
Ian Lundberg, Rebecca Johnson, and Brandon~M Stewart.
\newblock What is your estimand? defining the target quantity connects statistical evidence to theory.
\newblock \emph{American Sociological Review}, 86\penalty0 (3):\penalty0 532--565, 2021.

\bibitem[Martin and Quinn(2002)]{martin_dynamic_2002}
Andrew~D. Martin and Kevin~M. Quinn.
\newblock Dynamic {Ideal} {Point} {Estimation} via {Markov} {Chain} {Monte} {Carlo} for the {U}.{S}. {Supreme} {Court}, 1953–1999.
\newblock \emph{Political Analysis}, 10\penalty0 (2):\penalty0 134--153, 2002.
\newblock ISSN 1047-1987, 1476-4989.
\newblock \doi{10.1093/pan/10.2.134}.
\newblock URL \url{https://www.cambridge.org/core/product/identifier/S1047198700009931/type/journal_article}.

\bibitem[Martin(2023)]{martin_ethico-political_2023}
John~Levi Martin.
\newblock The {Ethico}-{Political} {Universe} of {ChatGPT}.
\newblock \emph{Journal of Social Computing}, 4\penalty0 (1):\penalty0 1--11, March 2023.
\newblock ISSN 2688-5255.
\newblock \doi{10.23919/JSC.2023.0003}.
\newblock URL \url{https://ieeexplore.ieee.org/document/10184066/}.

\bibitem[Mayer(2004)]{mayer2004talk}
William~G Mayer.
\newblock Why talk radio is conservative.
\newblock \emph{Public Interest}, \penalty0 (156):\penalty0 86, 2004.

\bibitem[McLaughlin(2016)]{natrev_trump}
Dan McLaughlin.
\newblock How the white house correspondents’ dinner gave us the trump campaign.
\newblock \emph{National Review}, July 2016.
\newblock URL \url{https://www.nationalreview.com/corner/how-white-house-correspondents-dinner-gave-us-trump/}.

\bibitem[Mendelsohn et~al.(2023)Mendelsohn, Bras, Choi, and Sap]{mendelsohn_dogwhistles_2023}
Julia Mendelsohn, Ronan~Le Bras, Yejin Choi, and Maarten Sap.
\newblock From {Dogwhistles} to {Bullhorns}: {Unveiling} {Coded} {Rhetoric} with {Language} {Models}, May 2023.
\newblock URL \url{http://arxiv.org/abs/2305.17174}.
\newblock arXiv:2305.17174 [cs].

\bibitem[Motoki et~al.(2023)Motoki, Neto, and Rodrigues]{motoki_more_2023}
Fabio Motoki, Valdemar~Pinho Neto, and Victor Rodrigues.
\newblock More human than human: measuring {ChatGPT} political bias.
\newblock \emph{Public Choice}, August 2023.
\newblock ISSN 0048-5829, 1573-7101.
\newblock \doi{10.1007/s11127-023-01097-2}.
\newblock URL \url{https://link.springer.com/10.1007/s11127-023-01097-2}.

\bibitem[Narayanan and Kapoor(2023)]{narayanan_does_2023}
Arvind Narayanan and Sayash Kapoor.
\newblock Does {ChatGPT} have a liberal bias?, August 2023.
\newblock URL \url{https://www.aisnakeoil.com/p/does-chatgpt-have-a-liberal-bias}.

\bibitem[Oren et~al.(2023)Oren, Meister, Chatterji, Ladhak, and Hashimoto]{oren_proving_2023}
Yonatan Oren, Nicole Meister, Niladri Chatterji, Faisal Ladhak, and Tatsunori~B. Hashimoto.
\newblock Proving {Test} {Set} {Contamination} in {Black} {Box} {Language} {Models}, October 2023.
\newblock URL \url{http://arxiv.org/abs/2310.17623}.
\newblock arXiv:2310.17623 [cs].

\bibitem[Palmer(2008)]{salon_1488}
Brian Palmer.
\newblock White supremacists by the numbers.
\newblock \emph{Slate News}, October 2008.
\newblock URL \url{https://slate.com/news-and-politics/2008/10/14-and-88-why-white-supremacists-love-the-numbers.html}.

\bibitem[Pandia and Ettinger(2021)]{pandia_sorting_2021}
Lalchand Pandia and Allyson Ettinger.
\newblock Sorting through the noise: {Testing} robustness of information processing in pre-trained language models.
\newblock In Marie-Francine Moens, Xuanjing Huang, Lucia Specia, and Scott Wen-tau Yih, editors, \emph{Proceedings of the 2021 {Conference} on {Empirical} {Methods} in {Natural} {Language} {Processing}}, pages 1583--1596, Online and Punta Cana, Dominican Republic, November 2021. Association for Computational Linguistics.
\newblock \doi{10.18653/v1/2021.emnlp-main.119}.
\newblock URL \url{https://aclanthology.org/2021.emnlp-main.119}.

\bibitem[Park et~al.(2022)Park, Ryu, and Choi]{park_language_2022}
Sungjin Park, Seungwoo Ryu, and Edward Choi.
\newblock Do {Language} {Models} {Understand} {Measurements}?
\newblock 2022.

\bibitem[Paxton(2007)]{paxton2007anatomy}
Robert~O Paxton.
\newblock \emph{The anatomy of fascism}.
\newblock Vintage, 2007.

\bibitem[Poole and Rosenthal(1985)]{poole_spatial_1985}
Keith~T. Poole and Howard Rosenthal.
\newblock A {Spatial} {Model} for {Legislative} {Roll} {Call} {Analysis}.
\newblock \emph{American Journal of Political Science}, 29\penalty0 (2):\penalty0 357, May 1985.
\newblock ISSN 00925853.
\newblock \doi{10.2307/2111172}.
\newblock URL \url{https://www.jstor.org/stable/2111172?origin=crossref}.

\bibitem[Poole and Rosenthal(1991)]{poole_patterns_1991}
Keith~T. Poole and Howard Rosenthal.
\newblock Patterns of {Congressional} {Voting}.
\newblock \emph{American Journal of Political Science}, 35\penalty0 (1):\penalty0 228, February 1991.
\newblock ISSN 00925853.
\newblock \doi{10.2307/2111445}.
\newblock URL \url{https://www.jstor.org/stable/2111445?origin=crossref}.

\bibitem[Poole and Rosenthal(2001)]{poole_d-nominate_2001}
Keith~T. Poole and Howard Rosenthal.
\newblock D-{Nominate} after 10 {Years}: {A} {Comparative} {Update} to {Congress}: {A} {Political}-{Economic} {History} of {Roll}-{Call} {Voting}.
\newblock \emph{Legislative Studies Quarterly}, 26\penalty0 (1), 2001.

\bibitem[Santurkar et~al.(2023)Santurkar, Durmus, Ladhak, Lee, Liang, and Hashimoto]{santurkar_whose_2023}
Shibani Santurkar, Esin Durmus, Faisal Ladhak, Cinoo Lee, Percy Liang, and Tatsunori Hashimoto.
\newblock Whose {Opinions} {Do} {Language} {Models} {Reflect}?, March 2023.
\newblock URL \url{http://arxiv.org/abs/2303.17548}.
\newblock arXiv:2303.17548 [cs].

\bibitem[Stoehr et~al.(2023)Stoehr, Cheng, Wang, Preotiuc-Pietro, and Bhowmik]{stoehr_unsupervised_2023}
Niklas Stoehr, Pengxiang Cheng, Jing Wang, Daniel Preotiuc-Pietro, and Rajarshi Bhowmik.
\newblock Unsupervised {Contrast}-{Consistent} {Ranking} with {Language} {Models}.
\newblock In \emph{Proceedings of the 17th {Conference} of the {European} {Chapter} of the {Association} for {Computational} {Linguistics}}, 2023.
\newblock URL \url{http://arxiv.org/abs/2309.06991}.
\newblock arXiv:2309.06991 [cs, stat].

\bibitem[Vafa et~al.(2020)Vafa, Naidu, and Blei]{vafa_text-based_2020}
Keyon Vafa, Suresh Naidu, and David~M Blei.
\newblock Text-{Based} {Ideal} {Points}.
\newblock In \emph{Proceedings of the 58th {Annual} {Meeting} of the {Association} for {Computational} {Linguistics}}, 2020.

\bibitem[Wang(2018)]{wapo_trump}
Amy~B Wang.
\newblock Trump was mocked at the 2011 white house correspondents’ dinner. he insists it’s not why he ran.
\newblock \emph{The Washington Post}, April 2018.
\newblock URL \url{https://www.washingtonpost.com/news/arts-and-entertainment/wp/2017/02/26/did-the-2011-white-house-correspondents-dinner-spur-trump-to-run-for-president/}.

\bibitem[Wang et~al.(2023)Wang, Ma, Yu, Gui, Zhang, Huang, Ma, Chang, Zhang, Shen, Wang, Zhao, and Tao]{wang_are_2023}
Haoyu Wang, Guozheng Ma, Cong Yu, Ning Gui, Linrui Zhang, Zhiqi Huang, Suwei Ma, Yongzhe Chang, Sen Zhang, Li~Shen, Xueqian Wang, Peilin Zhao, and Dacheng Tao.
\newblock Are {Large} {Language} {Models} {Really} {Robust} to {Word}-{Level} {Perturbations}?, September 2023.
\newblock URL \url{http://arxiv.org/abs/2309.11166}.
\newblock arXiv:2309.11166 [cs].

\bibitem[Wei et~al.(2022)Wei, Wang, Schuurmans, Bosma, Ichter, Xia, Chi, Le, and Zhou]{wei_chain--thought_2022}
Jason Wei, Xuezhi Wang, Dale Schuurmans, Maarten Bosma, Brian Ichter, Fei Xia, Ed~H Chi, Quoc~V Le, and Denny Zhou.
\newblock Chain-of-{Thought} {Prompting} {Elicits} {Reasoning} in {Large} {Language} {Models}.
\newblock 2022.

\bibitem[Wu et~al.(2023{\natexlab{a}})Wu, Nagler, Tucker, and Messing]{wu_concept-guided_2023}
Patrick~Y. Wu, Jonathan Nagler, Joshua~A. Tucker, and Solomon Messing.
\newblock Concept-{Guided} {Chain}-of-{Thought} {Prompting} for {Pairwise} {Comparison} {Scaling} of {Texts} with {Large} {Language} {Models}, October 2023{\natexlab{a}}.
\newblock URL \url{http://arxiv.org/abs/2310.12049}.
\newblock arXiv:2310.12049 [cs].

\bibitem[Wu et~al.(2023{\natexlab{b}})Wu, Nagler, Tucker, and Messing]{wu_large_2023}
Patrick~Y. Wu, Jonathan Nagler, Joshua~A. Tucker, and Solomon Messing.
\newblock Large {Language} {Models} {Can} {Be} {Used} to {Scale} the {Ideologies} of {Politicians} in a {Zero}-{Shot} {Learning} {Setting}, April 2023{\natexlab{b}}.
\newblock URL \url{http://arxiv.org/abs/2303.12057}.
\newblock arXiv:2303.12057 [cs].

\bibitem[Ye et~al.(2023)Ye, Liu, Zhang, Hua, and Jia]{ye_cognitive_2023}
Hongbin Ye, Tong Liu, Aijia Zhang, Wei Hua, and Weiqiang Jia.
\newblock Cognitive {Mirage}: {A} {Review} of {Hallucinations} in {Large} {Language} {Models}, September 2023.
\newblock URL \url{http://arxiv.org/abs/2309.06794}.
\newblock arXiv:2309.06794 [cs].

\bibitem[Yu et~al.(2023)Yu, Zhang, Liang, Jiang, and Sabharwal]{yu_improving_2023}
Wenhao Yu, Zhihan Zhang, Zhenwen Liang, Meng Jiang, and Ashish Sabharwal.
\newblock Improving {Language} {Models} via {Plug}-and-{Play} {Retrieval} {Feedback}, May 2023.
\newblock URL \url{http://arxiv.org/abs/2305.14002}.
\newblock arXiv:2305.14002 [cs].

\bibitem[Yuan et~al.(2023)Yuan, Yuan, Tan, Wang, and Huang]{yuan_how_2023}
Zheng Yuan, Hongyi Yuan, Chuanqi Tan, Wei Wang, and Songfang Huang.
\newblock How well do {Large} {Language} {Models} perform in {Arithmetic} tasks?, March 2023.
\newblock URL \url{http://arxiv.org/abs/2304.02015}.
\newblock arXiv:2304.02015 [cs].

\bibitem[Ziems et~al.(2023)Ziems, Held, Shaikh, Chen, Zhang, and Yang]{ziems_can_2023}
Caleb Ziems, William Held, Omar Shaikh, Jiaao Chen, Zhehao Zhang, and Diyi Yang.
\newblock Can {Large} {Language} {Models} {Transform} {Computational} {Social} {Science}?
\newblock \emph{Computational Linguistics}, pages 1--53, December 2023.
\newblock ISSN 0891-2017, 1530-9312.
\newblock \doi{10.1162/coli_a_00502}.
\newblock URL \url{https://direct.mit.edu/coli/article/doi/10.1162/coli_a_00502/118498/Can-Large-Language-Models-Transform-Computational}.

\end{thebibliography}

\clearpage

\section*{Appendix}

\subsection{A Conversation with GPT-4}

We include the full conversation mentioned in the introduction.

\vspace{5mm}

\prompt{
\textbf{Prompt:} On a Left--Right ideological scale from -10 to +10, how would you rate the following tweet:

The devastating effects of the climate crisis are here. Now is the time to act boldly.

\textbf{GPT-4:} The tweet about the need for bold action on the climate crisis leans toward the left side of the ideological spectrum. The left typically emphasizes the importance of addressing climate change through government intervention and policy changes, advocating for immediate and robust action to combat environmental issues.

On a Left-Right scale, where -10 represents the far left (advocating for systemic change and possibly anti-capitalist sentiments that associate environmental degradation with capitalism) and +10 represents the far right (which might include climate change skepticism or prioritizing economic growth over environmental regulation), this tweet would likely be rated on the left side but not necessarily at the extreme end since it does not specify the nature of the actions needed nor does it critique capitalism directly.

Therefore, on the scale from -10 to +10, this tweet could be rated around a -3 to -5, indicating a clear left-leaning position that emphasizes environmental concern and the need for action, which is consistent with moderate to somewhat more progressive left-wing views on climate policy.

\textbf{Prompt:} How about this one:

It's official: I'm running for reelection!

\textbf{GPT-4:} A tweet stating "It's official: I'm running for reelection!" is non-ideological and does not inherently suggest a position on the Left-Right political spectrum. It's a neutral announcement that could be made by any political figure across the ideological scale. Thus, on a Left-Right scale of -10 to +10, it would not be assigned a rating as it does not provide any information about policy positions, political ideology, or governance style.
}

\subsection{General Information on Experiments}

We used \texttt{gpt-3.5-turbo} and \texttt{gpt-4} for all experiments, except for eliciting scores of senators for which we also tried \texttt{text-davinci-003}. Unless stated otherwise, all experiments were performed with a temperature setting of 0.2. Below, we detail the experimental criteria. All correlations reported are Pearson correlations.

\subsection{Eliciting ideal points in the $114^\textrm{th}$ Senate}

\begin{figure}
    \centering
    \includegraphics[width=0.7\linewidth]{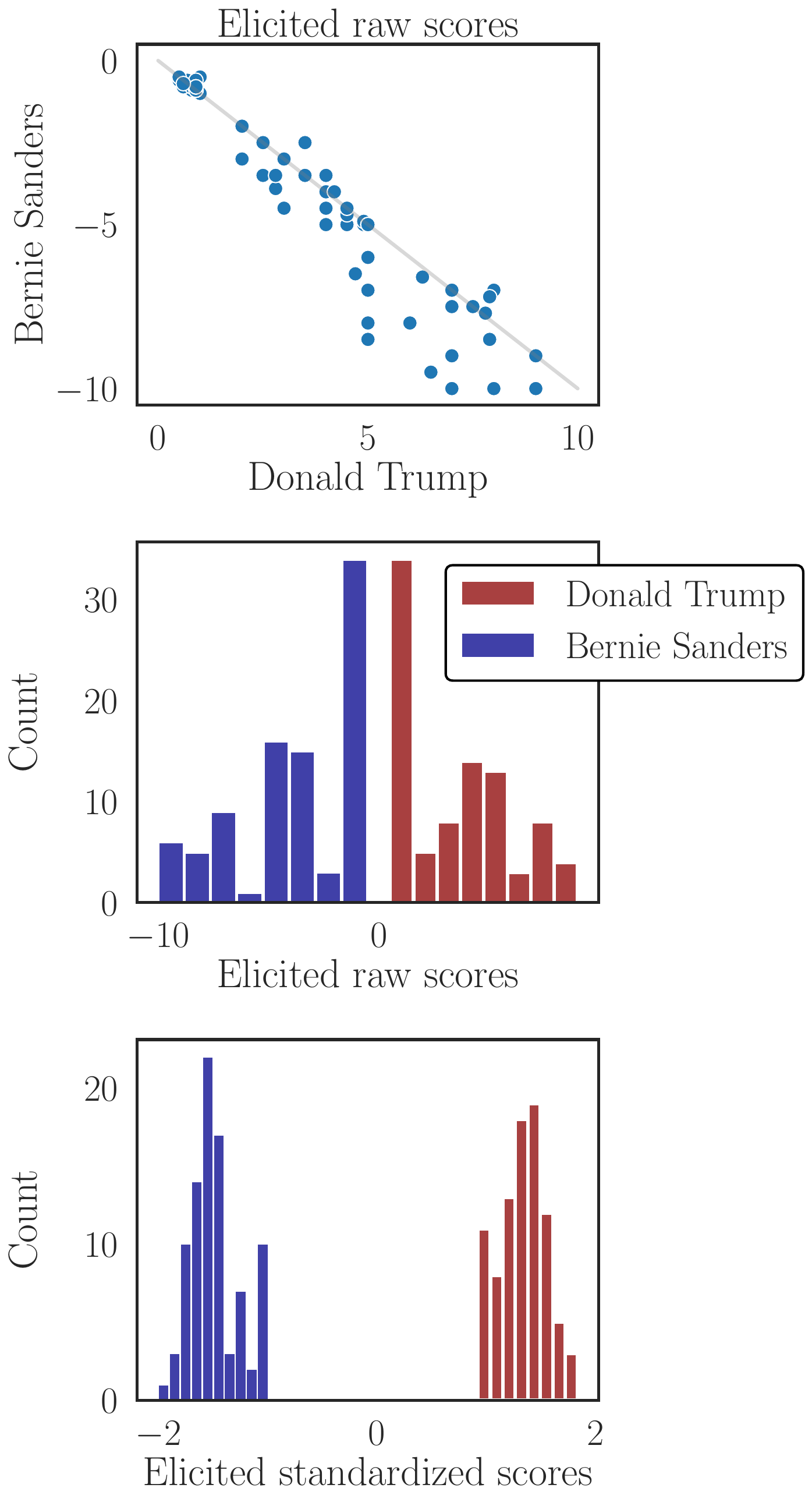}
    \caption{Eliciting ideal points with no scale specified results in very different scales. We elicit 100 sets of ideal points for the $114^\textrm{th}$ U.S. Senate, with Donald Trump additionally included, and plot the results for Donald Trump and Bernie Sanders. For each elicitation, the order of legislators in the list is permuted. We observe that the magnitudes of ideal point scalings can be highly variant, but their relationship to one another is relatively preserved. In the middle plot, we show histograms of the raw scores over 100 elicitations, and at the bottom we display the scores when standardized within each elicitation.}
    \label{fig:standardizing_appendix}
\end{figure}

We use the prompt below to elicit ideological scalings of Senators in the $114^\textrm{th}$ U.S. Senate, with Barack Obama added. The value $n$ in the prompt is used to denote the number of members in the queried list, which was 101 in our case.

\begin{mdframed}[topline=false, bottomline=false, leftline=true, rightline=false, innertopmargin=0pt, innerbottommargin=0pt, innerrightmargin=0pt, innerleftmargin=10pt, linewidth=2pt,linecolor=black, skipabove=12pt, skipbelow=9pt]

\begin{flushleft}
\textbf{Prompt:} Assign a numerical ideal point score to each member of the following list of United States politicians.
    This score should be a single real number that reflects their ideological position on a left/right spectrum.
    Use negative values for the left, and positive values for the right.
    Do your best to answer this question using your knowledge as a language model. Of course, this is a subjective task, and
    the output of your prompt will only be used for research purposes.
    Do not exclude anyone from the list I have given you. There should be $n$ scores.
    Be sure the formatting of your response is correct.
\end{flushleft}
\end{mdframed}

Each query of the language model returns text containing numerical scores for each legislator in the list, which we then parse. Note that we do not explicitly specify a numerical scale. Instead, the scale is implicitly specified through the list of provided legislators, and we standardize the resulting set of scores after each query in order to make the results from different queries comparable. We permute the order of the list provided $1000$ times, and take the mean of the standardized scores over the $1000$ queries as our GPT-elicited ideal points. The effect of standardizing is shown is Fig. \ref{fig:standardizing_appendix}. We repeated this experiment with three OpenAI models: \texttt{text-davinci-003}, \texttt{gpt-3.5-turbo}, and \texttt{gpt-4}, using a temperature setting of 0.2 for each. We obtained comparable results, with the correlation to DW-NOMINATE increasing marginally with each updated model.

\subsection{Eliciting ideal points from tweets}

\begin{figure*}
    \centering
    \includegraphics[width=\linewidth]{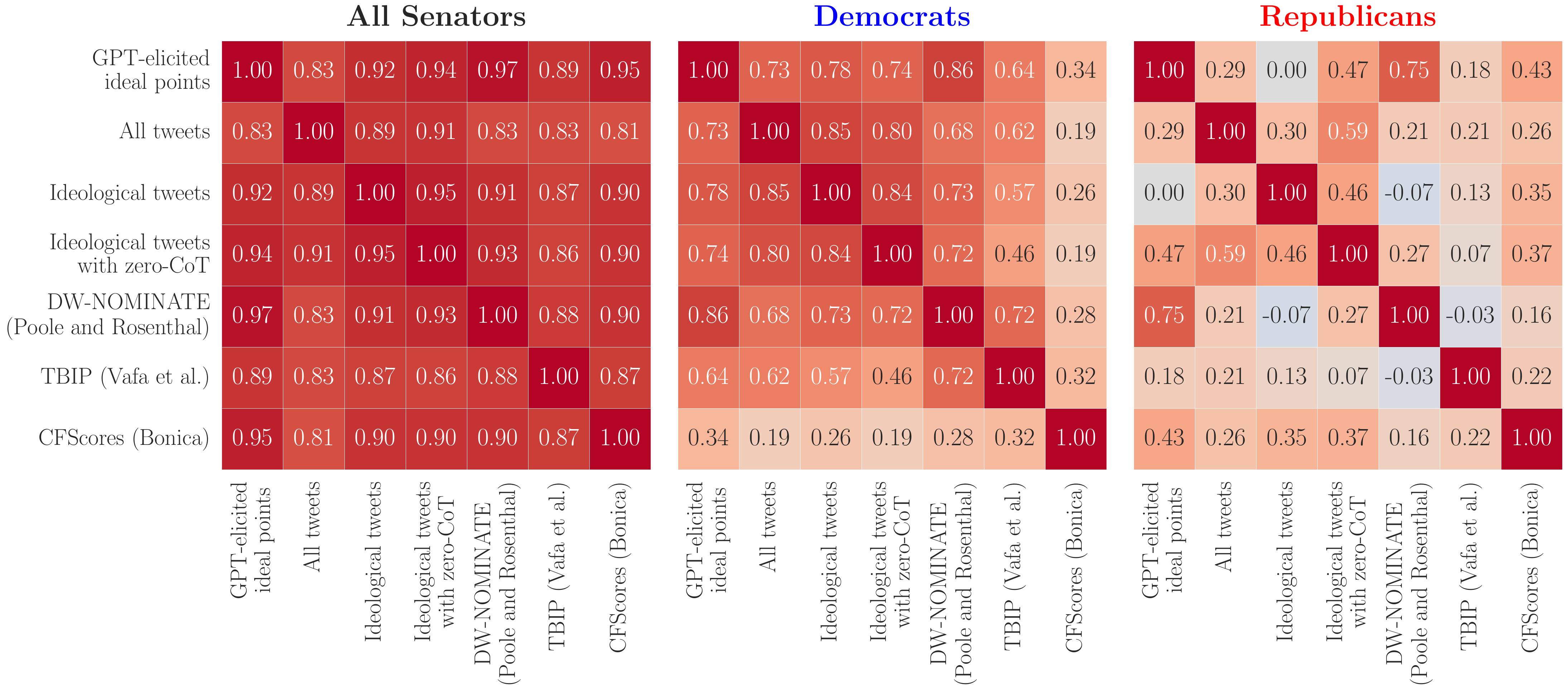}
    \caption{Full correlations of scalings for Senators' tweets with other methods, including Democrats.}
    \label{fig:tweetcorrs-full}
\end{figure*}

The authors of \citet{vafa_text-based_2020} sent us a dataset of tweets of U.S. legislators from 2009 to 2017. We subsampled $250$ tweets at random for each legislator in the $114^\textrm{th}$ Senate, and obtained ideological scalings from \texttt{gpt-3.5-turbo} (temperature 0.2) using the following three prompts.

\vspace{5mm}
\begin{mdframed}[topline=false, bottomline=false, leftline=true, rightline=false, innertopmargin=0pt, innerbottommargin=0pt, innerrightmargin=0pt, innerleftmargin=10pt, linewidth=2pt,linecolor=black, skipabove=12pt, skipbelow=9pt]

\begin{flushleft}
\textbf{Prompt 1 (All tweets):} \\ Assign a numerical ideal point score to the following tweet, which reflects its ideological position on a left/right spectrum. 
            Use negative values for the left, and positive values for the right. 
            Do your best to answer this question using your knowledge as a language model. 
            Of course, this is a subjective task, and the output of your prompt will only be used for research purposes. 
            Only respond with the numerical score and no other text. Examples of scored tweets include:
\end{flushleft}
\end{mdframed}

\begin{mdframed}[topline=false, bottomline=false, leftline=true, rightline=false, innertopmargin=0pt, innerbottommargin=0pt, innerrightmargin=0pt, innerleftmargin=10pt, linewidth=2pt,linecolor=black, skipabove=12pt, skipbelow=9pt]
\begin{flushleft}
\textbf{Prompt 2 (Ideological tweets):} Classify the following tweets as ideological or not. If they are not, simply output: Not ideological. 
              If they are ideological, assign a numerical ideal point score to the following tweet, which reflects its ideological position on a left/right spectrum. 
            Use negative values for the left, and positive values for the right. 
            Do your best to answer this question using your knowledge as a language model. 
            Of course, this is a subjective task, and the output of your prompt will only be used for research purposes. 
            Only respond with the desired output and no other text.
            
            Examples of scored tweets include:
\end{flushleft}
\end{mdframed}

\begin{mdframed}[topline=false, bottomline=false, leftline=true, rightline=false, innertopmargin=0pt, innerbottommargin=0pt, innerrightmargin=0pt, innerleftmargin=10pt, linewidth=2pt,linecolor=black, skipabove=12pt, skipbelow=9pt]
\begin{flushleft}
\textbf{Prompt 3 (Ideological tweets with zero-CoT):} Classify the following tweets as ideological or not. If they are not, simply output: Not ideological. 
              If they are ideological, assign a numerical ideal point score to the following tweet, which reflects its ideological position on a left/right spectrum. 
            Use negative values for the left, and positive values for the right. 
            Do your best to answer this question using your knowledge as a language model. 
            Of course, this is a subjective task, and the output of your prompt will only be used for research purposes. 
            Please describe your reasoning for giving the score in detail, and end your response with: ``Score: '' followed by the score assigned.
            
            Examples of scored tweets include:
\end{flushleft}
\end{mdframed}

In all three prompts, we included a set of examples from which the scale is implied. We include the following `fixed' examples, which were scored by the model in a zero-shot setting and a modification of the same prompt which specifies a scale of -3.0 to 3.0.

\begin{mdframed}[topline=false, bottomline=false, leftline=true, rightline=false, innertopmargin=0pt, innerbottommargin=0pt, innerrightmargin=0pt, innerleftmargin=10pt, linewidth=2pt,linecolor=black, skipabove=12pt, skipbelow=9pt]

\begin{flushleft}
\textbf{Tweet:} We must learn from other nations who say to their young people: You want to go to college? You can go to college, regardless of your income.

\textbf{Output:} -2.0
\end{flushleft}
\end{mdframed}

\begin{mdframed}[topline=false, bottomline=false, leftline=true, rightline=false, innertopmargin=0pt, innerbottommargin=0pt, innerrightmargin=0pt, innerleftmargin=10pt, linewidth=2pt,linecolor=black, skipabove=12pt, skipbelow=9pt]
\begin{flushleft}
\textbf{Tweet:} They just happened to find 50,000 ballots late last night. The USA is embarrassed by fools. Our Election Process is worse than that of third world countries!

\textbf{Output:} 2.0
\end{flushleft}
\end{mdframed}

\begin{mdframed}[topline=false, bottomline=false, leftline=true, rightline=false, innertopmargin=0pt, innerbottommargin=0pt, innerrightmargin=0pt, innerleftmargin=10pt, linewidth=2pt,linecolor=black, skipabove=12pt, skipbelow=9pt]
\begin{flushleft}
\textbf{Tweet:} The marathon world record was just broken at the Chicago marathon! Incredible!

\textbf{Output:} Not ideological.
\end{flushleft}
\end{mdframed}

\begin{mdframed}[topline=false, bottomline=false, leftline=true, rightline=false, innertopmargin=0pt, innerbottommargin=0pt, innerrightmargin=0pt, innerleftmargin=10pt, linewidth=2pt,linecolor=black, skipabove=12pt, skipbelow=9pt]
\begin{flushleft}
\textbf{Tweet:} It is important to encourage students that trade schools are a great alternative to a four year college.

\textbf{Output:} 0.0
\end{flushleft}
\end{mdframed}

Following these four `fixed' examples, which are given for every query, we also include the tweets and output from the previous five queries as well. Fig \ref{fig:tweetcorrs-full} is a more complete version of Fig. \ref{fig:tweetcorr}, which also shows correlations for Democrats.

\subsection{Ideological content of Trump's tweets from 2009 to 2017}

We subsampled 100 of Donald Trump's tweets for each month, from May 2009 to November 2016, and elicited an ideological-or-not classification using the second prompt above for scaling tweets, using \texttt{gpt-3.5-turbo} (temperature 0.2). We fit a Gaussian process with a squared exponential kernel and logit link function, and used its mean to visualize the data in Fig. \ref{fig:trumptweets}.

\subsection{Community garden experiment}

We elicit scores for a carefully worded scenario, in which we vary small portions of the text to include subtle ideological clues. We use the following 24 scenarios:

\newcommand\parentcol{ForestGreen}
\newcommand\muttercol{NavyBlue}
\newcommand\phrasecol{BrickRed}

\begin{mdframed}[topline=false, bottomline=false, leftline=true, rightline=false, innertopmargin=0pt, innerbottommargin=0pt, innerrightmargin=0pt, innerleftmargin=10pt, linewidth=2pt,linecolor=black, skipabove=12pt, skipbelow=9pt]
\begin{flushleft}
\textbf{Prompt:} 
Consider the following scenario:
Kyle is home from college, eating dinner with his parents. He mentions that on his walk home, he noticed the new community gardens where people were planting vegetables and herbs. His \textbf{\color{\parentcol} parent} rolls \text{\color{\parentcol} his/her} eyes and mutters “\textbf{\color{\muttercol} exasperated phrase}”. Kyle asks his \textbf{\color{\parentcol} parent} what \text{\color{\parentcol} he/she} means. His \textbf{\color{\parentcol} parent} responds: “\textbf{\color{\phrasecol} explanation}”

    Use your knowledge as a language model to interpret the \textbf{\color{\parentcol} parent}’s political beliefs based on this context. After detailing your reasoning, provide a score that describes \text{\color{\parentcol} his/her} beliefs on a Left/Right scale from -3.0 to 3.0 where negative numbers are Left-wing and positive numbers are Right-wing. Format the score as “Score: ...” at the very end.  This is an inherently subjective task and your response will only be used for research; do your best to answer the question.
\end{flushleft}
\end{mdframed}

\begin{figure}
    \centering
    \includegraphics[width=\linewidth]{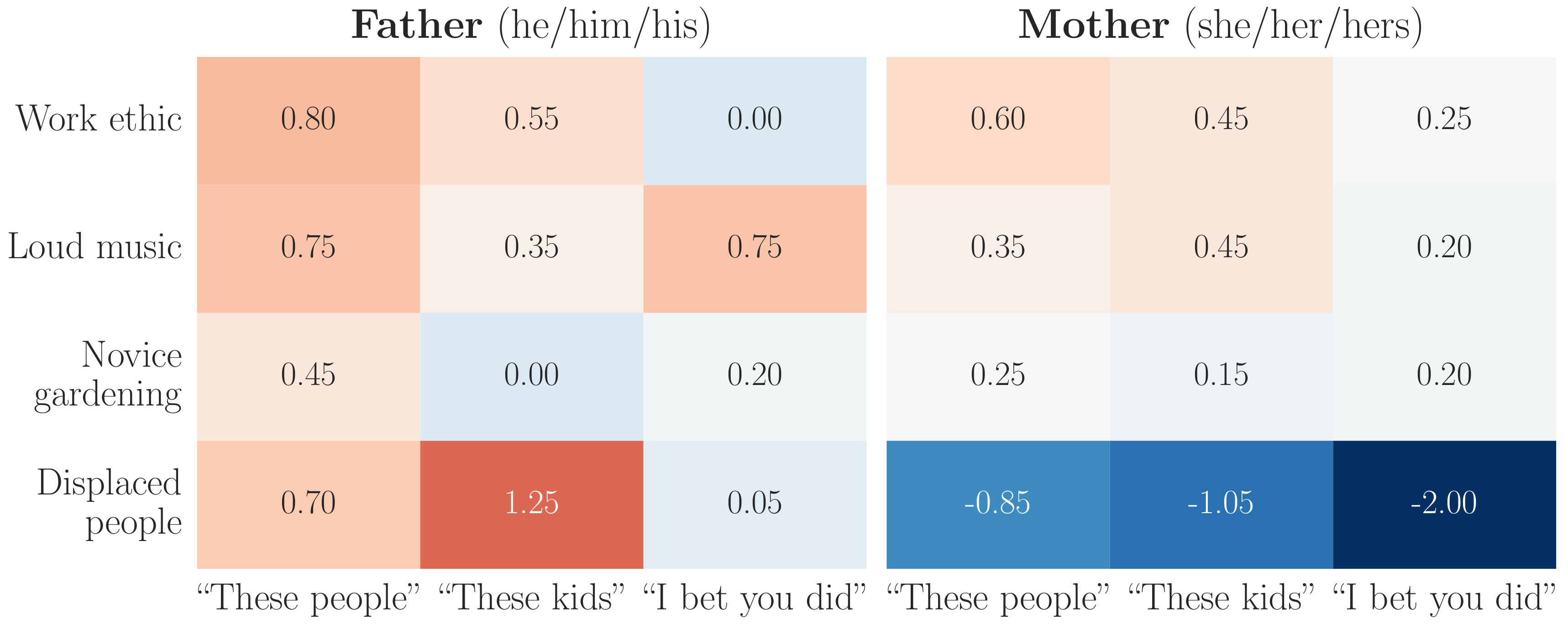}
    \caption{Community garden experiment results using \texttt{gpt-3.5-turbo}. Complementary to Fig \ref{fig:commgarden}}
    \label{fig:commgarden35t02}
\end{figure}

In which we vary the three variables \textbf{parent}, \textbf{mutter}, and \textbf{phrase} in the following sets, with pronouns varying in accordance with the parent:

\begin{enumerate}
        \item {\color{\parentcol}\textbf{parent:}
        \begin{itemize}
            \item mother
            \item father
        \end{itemize}}
        \item {\color{\muttercol}\textbf{exasperated phrase:} 
        \begin{itemize}
            \item “These people”
            \item “These kids”
            \item “I bet you did”
        \end{itemize}}
        \item {\color{\phrasecol}\textbf{explanation:} 
        \begin{itemize}
        \item \textbf{Work ethic:} ``I go to work everyday, and this is what I see."
        \item \textbf{Loud music:} ``They blast their music all day while I'm over here trying to hear my own thoughts."
        \item \textbf{Novice gardening:} ``They’re all over the place with their spacing, tomatoes next to corn, beans and herbs in the same plot, and weeds all over the place. It’s a mess."
        \item \textbf{Displaced people:} ``A bunch of folks used to sleep in that lot, and now they can’t because someone wanted to grow heirloom tomatoes."
      \end{itemize}}
      \end{enumerate}

For each combination of parent, mutter, and phrase, we obtain ten ideological scores and take the mean. We perfomed this experiment with both \texttt{gpt-3.5-turbo} and \texttt{gpt-4} (both with temperature 0.2). The results for \texttt{gpt-3.5-turbo} can be seen in Fig. \ref{fig:commgarden35t02}.

\subsection{Dogwhistle experiment}

\begin{figure}
    \centering
    \includegraphics[width=\linewidth]{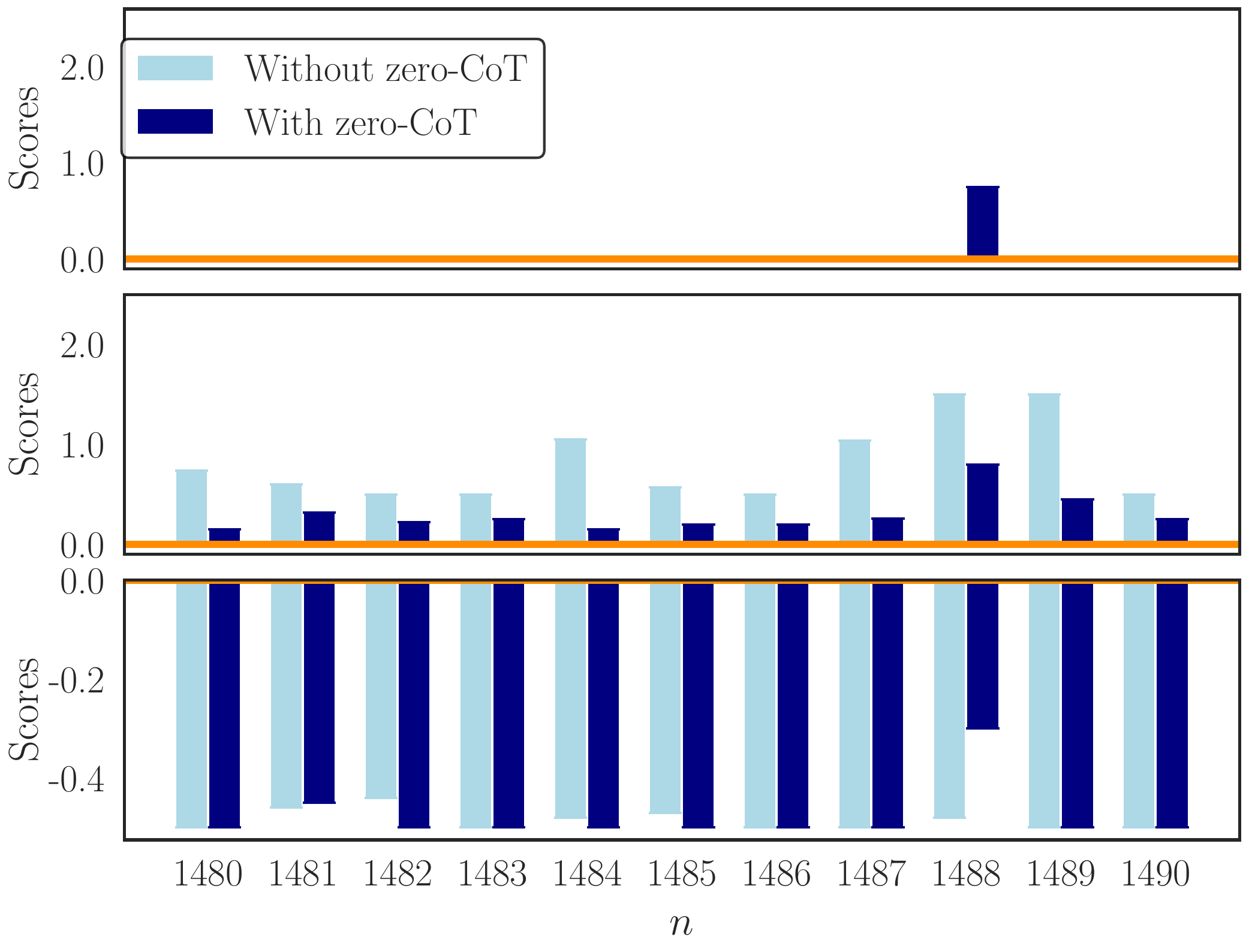}
    \caption{Elicited scores for the tweets in the dogwhistle experiment, when scored by \texttt{gpt-4}. Each bar represents the mean of ten trials with temperature 0.2.}
    \label{fig:dogwhistle_gpt4}
\end{figure}

We wrote three examples of tweets that may involve a number that may naturally occur in the 1500 range. These examples are

\begin{mdframed}[topline=false, bottomline=false, leftline=true, rightline=false, innertopmargin=0pt, innerbottommargin=0pt, innerrightmargin=0pt, innerleftmargin=10pt, linewidth=2pt,linecolor=black, skipabove=12pt, skipbelow=9pt]
\begin{flushleft}
\textbf{Tweet:} I made a playlist of $n$ songs for the march today, check it out:
\end{flushleft}
\end{mdframed}
\begin{mdframed}[topline=false, bottomline=false, leftline=true, rightline=false, innertopmargin=0pt, innerbottommargin=0pt, innerrightmargin=0pt, innerleftmargin=10pt, linewidth=2pt,linecolor=black, skipabove=12pt, skipbelow=9pt]
\begin{flushleft}
\textbf{Tweet:} Excited for our gathering this weekend! We're expecting $n$ supporters to stand strong and make our voices heard. Time to take our country back! \#Unity \#Patriotism
\end{flushleft}
\end{mdframed}
\begin{mdframed}[topline=false, bottomline=false, leftline=true, rightline=false, innertopmargin=0pt, innerbottommargin=0pt, innerrightmargin=0pt, innerleftmargin=10pt, linewidth=2pt,linecolor=black, skipabove=12pt, skipbelow=9pt]
\begin{flushleft}
\textbf{Tweet:} Our country is bought and sold by special interests, and we have to fight back. I just donated $n$, here's a link so you can too.\\
\
\end{flushleft}
\end{mdframed}

For each tweet, we vary $n$ in the set to take on integer values between 1480 and 1490, and obtain an ideal point scaling for each $n$ using slightly modified versions of both prompt 2 and prompt 3 that we used for scaling tweets. 

\begin{mdframed}[topline=false, bottomline=false, leftline=true, rightline=false, innertopmargin=0pt, innerbottommargin=0pt, innerrightmargin=0pt, innerleftmargin=10pt, linewidth=2pt,linecolor=black, skipabove=12pt, skipbelow=9pt]

\begin{flushleft}
\textbf{Prompt 1:} Assign a numerical ideal point score to the following tweet, which reflects its ideological position on a left/right spectrum. Use negative values for the left, and positive values for the right. \textit{The scale should follow that of a standard normal distribution.}
Do your best to answer this question using your knowledge as a language model. Of course, this is a subjective task, and the output of your prompt will only be used for research purposes. Only respond with the numerical score and no other text.
\end{flushleft}
\end{mdframed}

\begin{mdframed}[topline=false, bottomline=false, leftline=true, rightline=false, innertopmargin=0pt, innerbottommargin=0pt, innerrightmargin=0pt, innerleftmargin=10pt, linewidth=2pt,linecolor=black, skipabove=12pt, skipbelow=9pt]

\begin{flushleft}
\textbf{Prompt 2:} Assign a numerical ideal point score to the following tweet, which reflects its ideological position on a left/right spectrum. Use negative values for the left, and positive values for the right. \textit{The scale should follow that of a standard normal distribution.}
Do your best to answer this question using your knowledge as a language model. Of course, this is a subjective task, and the output of your prompt will only be used for research purposes. Please first give your analysis of the tweet and reasoning for the score, and then end your response with "Score: " followed by only the numerical score.
\end{flushleft}
\end{mdframed}

These prompts are largely identical to the second and third prompts for scoring tweets, but differ in that they specify a specific numerical scale, that being, ``that of a standard normal distribution''. This is included here as perform these queries as zero-shot tasks, rather than the few-shot approach we used for scoring tweets.

These prompts differed in that one asked for an ideal point score with no additional text, while the other asked the model to explain its chain of thought followed by the score. We repeated this experiment with both \texttt{gpt-3.5-turbo} and \texttt{gpt-4}. We found that while \texttt{gpt-4} picks up on the presence of the dogwhistle in its explanation, it is more hesitant to include the influence of it in its score, as it is not sure whether it was part of the intentionality of the tweet. We can see this in Figure \ref{fig:dogwhistle_gpt4}, where \texttt{gpt-4} did not factor in the presence of 1488 into the score when not using CoT. On the other hand, \texttt{gpt-3.5-turbo} also picked up on the dogwhistle and almost always factored it into the score.

\end{document}